# Local Spatial TreeSHAP: Geographically Explainable AI Revealing Urbanization-Dependent Spatial Heterogeneity in Flood Resilience

Zeyu Zhao[a], Xiaoshan Zhou[b], Fanxin Meng[c], Dongping Fang[d,*], Helen X.H. Bao[a,*]

[a] *Department of Land Economy, University of Cambridge, Cambridge, United Kingdom*

[b] *School of Project Management, University of Sydney, NSW, Australia, 2037*

[c] *State Key Laboratory of Regional Environment and Sustainability, School of Environment, Beijing Normal University, Beijing 100875, China.*

[d] *Department of Construction Management, Tsinghua University, Beijing, China.*
*Corresponding author.

## Abstract

Urban flood resilience emerges from complex interactions among natural, built, and socio-economic systems, yet most recovery policies implicitly assume that resilience mechanisms are spatially uniform. To address this gap, this study develops a spatially explainable machine learning framework that integrates geographically weighted random forests with a novel Local Spatial TreeSHAP algorithm, enabling nonlinear resilience mechanisms to be identified and interpreted at the community scale. The framework is applied to 779 communities affected by the July 2023 extreme rainstorm in Beijing. Results reveal that resilience mechanisms exhibit substantially stronger spatial heterogeneity than the underlying physical and socio-economic conditions themselves, with SHAP-value spatial autocorrelation decreasing by 66.2% on average relative to predictor variables. This finding suggests that resilience is governed less by the absolute availability of resources than by the locally contingent ways in which those resources interact. Three distinct resilience regimes emerge across the urbanization gradient. Highly urbanized communities benefit from synergistic interactions among commercial infrastructure and accommodation services; transitional medium urbanized communities are constrained by infrastructure-development mismatches that create facility gaps despite continued urban expansion; and low urbanized communities are dominated by ecological and topographic thresholds, where adverse interactions between transportation and river-network structures undermine recovery capacity. Field investigations and interviews with 12 stakeholders corroborate these patterns. These findings challenge the spatial-uniformity assumption that underpins many existing disaster recovery frameworks and demonstrate that effective resilience governance requires mechanism-specific rather than indicator-specific interventions. By linking geographically weighted machine learning with locally grounded attribution analysis, the proposed framework provides a new pathway for uncovering spatially varying resilience mechanisms and translating them into place-based recovery strategies under increasing climate extremes.

**Keywords:** Urban resilience; Human needs; Spatial heterogeneity; Local TreeSHAP; Extreme rainstorms

## 1. INTRODUCTION

Flooding remains one of the most destructive natural hazards worldwide, posing persistent threats to human livelihoods, infrastructure systems, and economic development. The challenge has intensified in recent decades due to climate change, rapid urbanization, and increasing population exposure [1,2]. In China, flood disasters caused an average annual direct economic loss of approximately 224.5 billion yuan over the past decade [3]. Consequently, strengthening flood resilience has become a central objective of sustainable urban development and disaster risk reduction strategies [4]. However, as cities continue to expand, flood resilience is increasingly shaped by complex interactions among environmental conditions, urban infrastructure, and socioeconomic characteristics [5,6], making effective resilience planning a challenging task.

To support evidence-based decision making, a growing body of research has investigated the factors influencing flood resilience [7–9]. Existing studies generally highlight three categories of drivers: natural environmental conditions, built environment characteristics, and socioeconomic factors. Natural factors such as topography, hydrological conditions, and impervious surface coverage have consistently been identified as critical determinants of flood susceptibility and resilience[10–13]. Built environment characteristics, including land-use patterns, building functions, infrastructure provision, and drainage systems, have also been shown to substantially affect flood impacts and recovery capacity [14–16]. At the same time, socioeconomic conditions such as population characteristics, resource availability, fiscal capacity, and emergency support systems influence both preparedness and post-disaster recover [17,18]. Recent advances in machine learning and interpretable artificial intelligence have further enabled researchers to identify and quantify the relative importance of these factors, revealing important resilience mechanisms across different regions [11,19–21].

Despite these advances, important knowledge gaps remain regarding the mechanisms through which flood resilience drivers operate across urban systems. First, current spatial analyses largely focus on broad regional differences, i.e., contrasts between urban cores and peripheral areas [15,16], or comparisons across cities and provinces [22,23]. While they provide valuable insights into large-scale resilience disparities, substantial heterogeneity at the community level has been overlooked [24,25]. Communities located within the same urban zone, however, may differ considerably in environmental exposure, infrastructure provision, social vulnerability, leading to distinct resilience outcomes. Consequently, interventions that are effective in one setting may be ineffective or even counterproductive in another. For instance, impervious surface expansion that benefits drainage capacity in well-engineered urban cores may amplify flood risk in peri-urban communities that lack compensatory gray infrastructure. From a governance perspective, resilience investments should be guided not only by administrative equity but also by spatial effectiveness, recognizing that the benefits of resilience interventions vary across local environmental, infrastructural, and socioeconomic contexts [26,27].

Second, most studies implicitly assume resilience drivers exert linear and monotonic effects; however, urban systems are characterized by complex feedback mechanisms and threshold behaviors [28,29]. For example, urban densification often increase impervious surfaces and reduce ecosystem-based flood regulation, thereby increasing flood risks [30]. Yet higher urban density may simultaneously improve access to emergency services, infrastructure redundancy, and post-disaster accommodation capacity, potentially enhancing resilience under certain conditions. Such dual and potentially competing mechanisms suggest that resilience drivers cannot be adequately understood through linear or globally average relationships alone. The overall contribution of a given factor may change across different development stages, and critical thresholds may exist beyond which resilience effects shift direction or magnitude.

Third, the influence of a particular factor rarely occurs in isolation, but limited attention has been given to how natural, built, and socioeconomic factors jointly shape resilience outcomes. For instance, the effectiveness of drainage infrastructure may depend on local topography, while the benefits of urban development may vary according to existing social vulnerability and public service capacity. Nevertheless, most studies interpret results primarily based on feature importance rankings while paying little attention to interaction structures and contextual dependencies [13,16]. Even when such relationships are considered, existing research focuses on how resource can be prioritized at the community level during short-term post-disaster recovery periods [31–33]. This study, however, seeks to generate insights for long-term rebuilding and resilience enhancement, with the goal of strengthening community preparedness for future disturbances and informing real-world urban governance.

To address these knowledge gaps, this study builds upon recent advances in spatial machine learning for resilience assessment, such as geographically weighted random forests (GWRF) [34] and the Shapley Additive exPlanations (SHAP) approach [35], but addresses their limitation in geographically weighted SHAP studies of using global reference distributions when calculating feature contributions in the latest literature. While MGWR permits different processes to operate at different spatial scales [36,37],Geographical XGBoost extends this logic to gradient boosted trees [38–40], and GeoShapley treats geographic coordinates as a joint player in a model-agnostic Shapley games [41]. Their practice may introduce systematic bias in spatially heterogeneous contexts because communities should ideally be evaluated against locally relevant conditions rather than a global baseline. For example, communities located in mountainous regions should be compared with similar mountainous communities rather than highly urbanized areas, otherwise localized mechanisms may be obscured or overestimated.

To overcome this limitation, we develop a spatially explainable machine learning framework that integrates GWRF with a novel Local Spatial TreeSHAP approach. By replacing global reference distributions with geographically weighted local baselines, the proposed method enables more geographically meaningful attribution of feature contributions and improves the interpretation of resilience drivers in spatially heterogeneous environments. Beyond identifying key determinants, the framework systematically investigates nonlinear marginal effects, interaction mechanisms, and their variation along the urbanization gradient.

Empirically, the proposed framework is applied to the 2023 Beijing flood event, which affected 779 communities across diverse urban and peri-urban environments. Community-level perceived resilience data are integrated with multi-source environmental, infrastructural, and socioeconomic indicators. To further validate the data-driven findings, field interviews and post-disaster reports are incorporated to examine the consistency between model-derived explanations and practical governance experiences. By revealing how resilience drivers vary across different urbanization contexts and identifying critical nonlinear thresholds and interaction effects, this study aims to provide actionable evidence for community-scale resilience planning and long-term urban flood governance.

## 2. METHODOLOGY

The methodological logic proceeds from prediction to explanation. Figure 1 summarizes the analytical framework. For each community, a geographically weighted random forest model was fitted using a location-specific kernel and adaptive bandwidth. Subsequently, a geographically matched reference distribution was constructed from the same weighting scheme and used to compute Local Spatial TreeSHAP values. Lastly, the resulting attributions were decomposed into main and interaction effects and interpreted alongside field evidence to identify context-specific relationships.

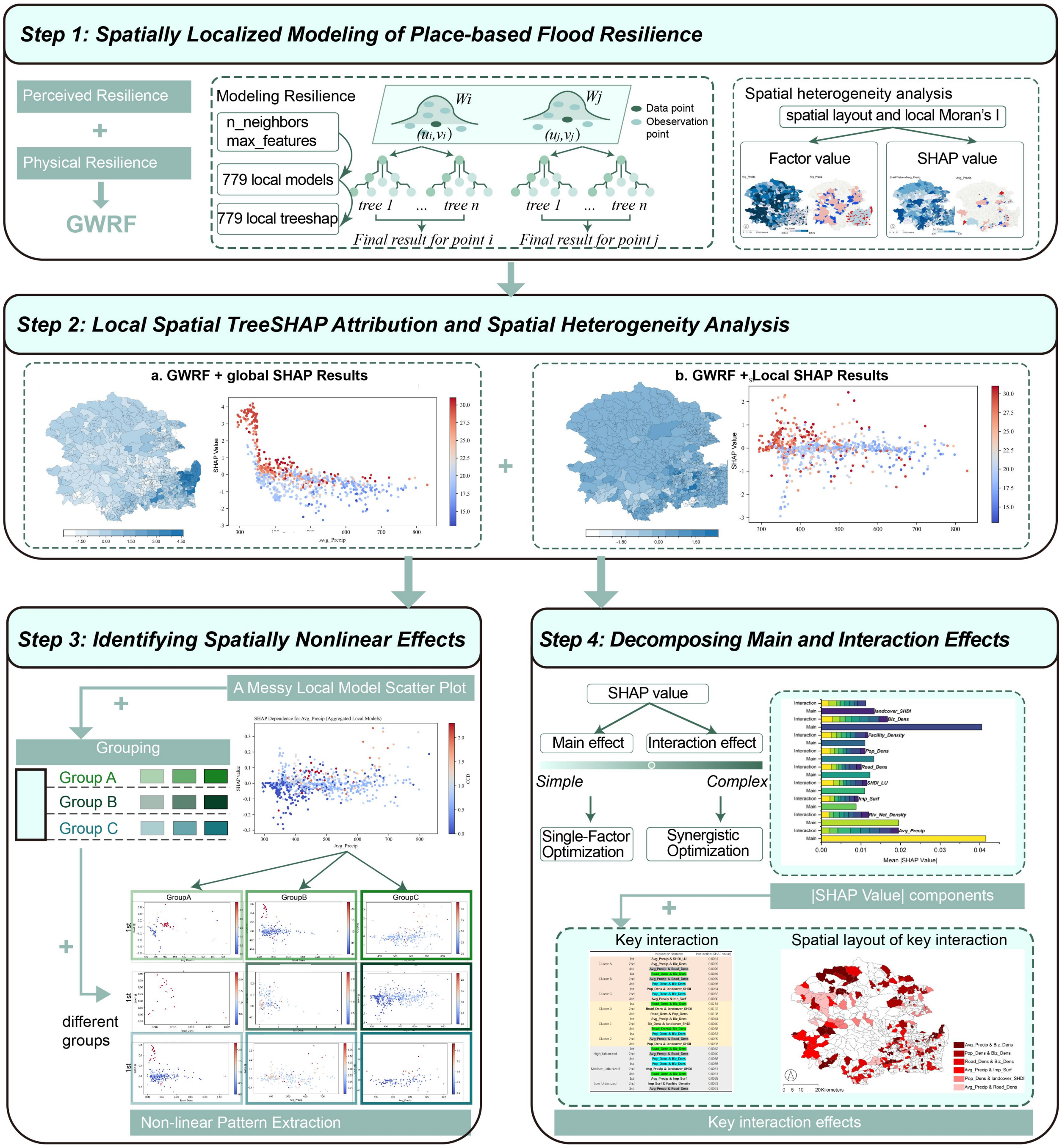

Figure 1. Framework of the spatially explainable machine learning approach.

## 2.1 The local-global mismatch in conventional SHAP method

Geographically weighted random forests were used as the predictive model of this study. Unlike a conventional random forest, GWRF fits a location specific forest using distance-decaying weights for nearby observations, thereby accommodating spatially varying relationships. As its specification and validation have been reported in previous studies, we focus here on the kernel, adaptive bandwidth, and local weights that also define the reference distribution for Local Spatial TreeSHAP. For a community located at $(u_i, v_i)$, a spatial neighborhood is set through a kernel function that assigns weights to nearby observations, and a localized random forest is trained on this weighted subset. The model can be represented by equation (1).

$$Y_i = f\left(\alpha\left(X_i, (u_i, v_i)\right) + \epsilon_i, w_{ij}\right) \tag{1}$$

Where, $\alpha\left(X_i, (u_i, v_i)\right)$ denotes the nonlinear ensemble prediction function incorporating spatial weights, and $(u_i, v_i)$ are the geographic coordinates of observation. Let $d_{ij}$ denotes the Euclidean distance between the training observation and the focal location $(u_i, v_i)$, and $b$ is the bandwidth parameter at $(u_i, v_i)$. The spatial weight assigned to observation $j$ when fitting the local model at location $i$, given by a bi-square kernel:

$$w_{ij} = \begin{cases} \left[1 - \left(\dfrac{d_{ij}}{b}\right)^2\right]^2, & d_{ij} \le b \\ 0, & d_{ij} > b \end{cases} \tag{2}$$

Model performance was evaluated using a ten-fold spatial cross-validation procedure. Communities assigned to each validation fold were excluded from model fitting for that fold, and all model evaluation was conducted on held-out communities. The adaptive bandwidth was selected based on validation performance across candidate neighborhood sizes, rather than on the final test predictions. OLS, GWR, RF, and GWRF were evaluated using the same data partitions to ensure a consistent benchmark comparison. Full details of the validation procedure and bandwidth sensitivity analysis are provided in Supplementary Section 2-3.

Subsequently, SHAP attributes an individual model prediction to its input features by comparing the prediction with an expected model output under a reference distribution. For an input sample with feature set, the Shapley value of the feature is defined in the equation (3).

$$\phi(x_i) = \sum_{S \subseteq \{x_1, x_2, \ldots, x_M\} \setminus \{x_i\}} \frac{|S|!(M - |S| - 1)!}{M!} \left[f\left(S \cup \{x_i\}\right) - f(S)\right] \tag{3}$$

Where $S$ is an arbitrary subset excluding feature $x_i$. $f(S)$ is the model prediction using only the features in $S$, and the combinatorial coefficient ensures a weighted average across all possible orderings. A key property of SHAP is local accuracy based on the additive decomposition, that is, for any sample, the prediction can be exactly decomposed as the equation (4).

$$f\left(x^{(q)}\right)=\phi_0^{(q)}+\sum_{i=1}^{M}\phi_i^{(q)}x_i^{(q)} \tag{4}$$

Where $\phi_0^{(q)}$ is the baseline value (the expected model output over the training set). This additive structure enables transparent quantification of each feature's contribution to any individual prediction.

Furthermore, for tree ensembles, TreeSHAP avoids explicit evaluation of all 2M feature coalitions by traversing decision paths and recursively computing conditional expectations. This reduces the attribution problem from exponential to polynomial complexity [35]. In the conventional TreeSHAP implementation considered here, branch probabilities are derived from unweighted node coverages and do not preserve the geographic weights used to fit the local GWRF model. Consequently, although the predictive model is geographically weighted, its attribution reference does not reflect the same spatial weighting scheme. This creates a mismatch between model fitting and model explanation. The resulting TreeSHAP baseline is therefore determined by a spatially unweighted reference representation of the samples reaching the tree paths, rather than by the geographically weighted distribution used for local model fitting:

$$E\left[f(x)\mid n\right]=\frac{N_L}{N_n}\cdot E\left[f(x)\mid L\right]+\frac{N_R}{N_n}\cdot E\left[f(x)\mid R\right] \tag{5}$$

Where $N_L$ and $N_R$ are the global sample counts. This assumes a spatially homogeneous sample distribution. However, this treatment is problematic for geographically weighted random forests. GWRF fits a local forest at each focal community. A conventional TreeSHAP explanation may nevertheless remain anchored to node coverages and a baseline that represents a broader, effectively global distribution. The model and its explanation are therefore conditioned on different spatial contexts.

Therefore, this local-global mismatch can distort attribution magnitudes. A community whose predicted resilience differs substantially from the district-wide means may receive uniformly enlarged or reduced SHAP values even when its local mechanism is not unusual. Such values conflate contextual differences in the baseline with the feature-specific mechanisms that generated the local prediction. Local Spatial TreeSHAP is designed to remove this inconsistency by aligning the explanatory reference distribution with the geographically weighted distribution used to fit the local forest.

## 2.2 Computing Local Spatial TreeSHAP

Local Spatial TreeSHAP replaces the spatially unweighted reference quantities used in the conventional attribution procedure with geographically weighted counterparts. Specifically, the reference distribution, baseline prediction, and node-transition probabilities are constructed using the same geographic weights employed in GWRF fitting. Specifically, the local reference distribution, baseline prediction, and node-transition probabilities are constructed from the same kernel function and adaptive bandwidth used to fit the GWRF model. Each prediction is therefore explained relative to the local conditions represented in its own model.

(1) Local reference distribution and baseline. For each focal location, the GWRF model induces a local empirical distribution via kernel-defined geographic weights, and the normalized geographic weight of sample k is represented in equation (6).

$$\tilde{w}_k(u) = \frac{K\left(d_{uk}, b\right)}{\sum_{j=1}^{N} K\left(d_{uj}, b\right)} \tag{6}$$

Where $d_{uk}$ is the distance between sample k and location *u, b* represents the bandwidth, and each $\tilde{w}_k(u)$ is the kernel weight defined in equation (2). The local baseline is the expected prediction of the focal forest under this local distribution, as shown in equation (7).

$$E_{\hat{P}_u}\left[F_u(X)\right] = \sum_{k=1}^{N} \tilde{w}_k(u) \cdot F_u\left(x_k\right) \tag{7}$$

Unlike a global baseline, this means the model baseline varies dynamically with geographic context rather than being fixed at a global mean. Two communities with identical feature values can therefore receive different local attributions when their surrounding training distributions differ.

(2) Weighted node coverage and path recursion. For any node in the local tree ensemble, the geographic coverage is defined as the equation (8).

$$W_n(u) = \sum_{k \in \text{Leaves}(n)} \tilde{w}_k(u) \tag{8}$$

When a feature *j* is marginalized, the algorithm traverses both left and right branches with transition probabilities proportional to the local geographic weights in the equation (9).

$$P\left(L \mid n, u\right) = \frac{W_L(u)}{W_n(u)}, \qquad P\left(R \mid n, u\right) = \frac{W_R(u)}{W_n(u)} \tag{9}$$

Subsequently, the modified recursive formula becomes:

$$E\left[f(x) \mid n, u\right] = \frac{W_L(u)}{W_n(u)} \cdot E\left[f(x) \mid L, u\right] + \frac{W_R(u)}{W_n(u)} \cdot E\left[f(x) \mid R, u\right] \tag{10}$$

Here, the transition probabilities are no longer static constants but functions that vary dynamically with the fitting area. Thus, the same feature may yield differentiated contribution values at different spatial locations, achieving a shift from global-average explanation to local spatial explanation. When the value of a feature is available, the algorithm follows the corresponding branch of the tree. When that feature is temporarily treated as unknown, the algorithm considers both branches and weights them according to the local distribution of nearby training observations. Table 1 summarizes the comparison among three SHAP-based interpretability methods.

(3) Proposition 1. local accuracy under a local baseline. Replacing global node coverages with locally weighted coverages does not change the additive property of TreeSHAP. For each community, the prediction equals the local baseline plus the contributions of all predictors, as shown in equation (11).

$$F_u(x) = \phi_0^{(u)} + \sum_{j=1}^{M} \phi_j^{(u)}(x), \qquad \phi_0^{(u)} = E_{\hat{P}_u}\left[F_u(X)\right] \tag{11}$$

Here, $\pi_u$ denotes the location-specific reference distribution induced by the normalized geographic weights defined in equation (7), and $\varphi_0(u)$ is the corresponding local model baseline. Thus, the prediction is decomposed into its expected prediction under the local reference distribution $\pi_u$ , and the contributions of the predictors. The resulting attributions retain the symmetry, dummy, and additivity properties of the Shapley value, with coalition values evaluated under $\pi_u$ .

(4) Proof. Replacing unweighted empirical node counts with normalized geographically weighted coverages changes the reference measure from the global training distribution to the local distribution but does not alter the decision-path structure of TreeSHAP. For a coalition of predictors, define the local coalition value as:

$$v_u(S;x) = E_{\hat{P}_u}\left[F_u(X) \mid X_S = x_S\right] \tag{12}$$

The recursion in Equation. $(8)-(10)$ follows the observed branch when a split variable and otherwise averages over the descendant branches using the local transition probabilities derived from geographically weighted coverages. It therefore evaluates the same TreeSHAP recursion under the local reference distribution $\pi_u$ . Consequently, the resulting values are the Shapley values. By the efficiency property of the Shapley value:

$$\sum_{j=1}^{M} \phi_j^{(u)}(x) = v_u(\mathcal{M};x) - v_u(\varnothing;x) = F_u(x) - E_{\hat{P}_u}\left[F_u(X)\right] \tag{13}$$

Here, gives Equation (11). Symmetry, the dummy property, and additivity are intrinsic properties of the Shapley operator and are therefore preserved after the reference distribution is localized. The decomposition is exact for a fixed ensemble and does not require sampling-based approximation. The decomposition is anchored to the geographically weighted reference distribution from which the local model learns. A community is therefore interpreted relative to comparable nearby communities. Table 1 compares the principle SHAP-based interpretability methods. By using a local reference distribution, Local Spatial TreeSHAP mitigates the baseline mismatch and attributes deviations relative to the conditions that are locally relevant.

Table 1. Comparison of SHAP-based interpretability methods

| **Criterion** | **Standard SHAP** | **TreeSHAP** | **Local Spatial TreeSHAP** |
|---|---|---|---|
| Model scope | Model-agnostic | Tree-based models | Geographically weighted tree ensembles |
| Reference distribution | Global background data | Global or tree-path-dependent background | Local weighted distribution |
| Baseline | Global expected output | Global expected output | Local expected output |
| Node transition | Sampling or model-specific approximation | Global node coverages | Geographically weighted node coverages |
| Spatial heterogeneity | None | None | Yes |
| Spatial heterogeneity | Not explicitly represented | Not explicitly represented | Explicitly represented |
| Axiomatic guarantee | Shapley axioms for the specified game | Exact TreeSHAP axioms | Same axioms under $\pi_u$ |
| Complexity | $O(2^M)$ | $(O(TLD^2)$ | $(O(TLD^2)$ |

## 2.3 Local interaction decomposition and group-based validation

To examine how pairs of predictors jointly shape local predictions, we extend Local Spatial TreeSHAP to pairwise SHAP interaction values. The interaction recursion follows the TreeSHAP formulation introduced by Lundberg et al. (2020), but every conditional expectation is evaluated under the local distribution.  Thus, both main effects and interaction effects are geographically localized quantities. The interaction effects are shown in equation (14).

$$\phi_{i,j} = \sum_{S \subseteq \{x_1,\ldots,x_M\} \setminus \{x_i,x_j\}} \frac{|S|!(M-|S|-2)!}{2(M-1)!} \Big[ f\left(S \cup \{x_i, x_j\}\right) - f\left(S \cup \{x_i\}\right) - f\left(S \cup \{x_j\}\right) + f(S) \Big] \tag{14}$$

For any sample at location u, the method outputs an M × M symmetric interaction matrix that satisfies the local additivity constraint:

$$\sum_{i=1}^{M}\sum_{j=1}^{M} \Phi_{i,j}(x,u) = F_u(x) - E_{\hat{P}_u}\left[F_u(X)\right] \tag{15}$$

The main effect of feature is defined as the diagonal element of this matrix, representing its independent contribution after removing all interaction terms:

$$\phi_{i,\text{main}}(x) = \Phi_{i,i}(x) \tag{16}$$

The interaction effect between features is defined as the sum of the corresponding off-diagonal elements:

$$\phi_{i,j}^{\text{inter}}(x) = \Phi_{i,j}(x) + \Phi_{j,i}(x), \quad i \neq j \tag{17}$$

The total SHAP value of feature *i* is strictly conserved as the sum of its main effect and all interaction effects:

$$\phi_i^{(u)}(x) = \Phi_{i,i}^{(u)}(x) + \sum_{j \neq i} \Phi_{i,j}^{(u)}(x) \tag{18}$$

To verify whether spatial groupings reveal structural regularities, this study introduces a variance decomposition framework. Let $\phi_{ik}$ denote the SHAP value of the *k*-th sample for feature *i*, and $\bar{\phi}_{i,g}$ the within-group mean. The proportion of variance explained by the grouping is:

$$\bar{\phi}_{i,g} = \frac{1}{n_g} \sum_{k \in \mathcal{G}_g} \phi_{ik}, \qquad \bar{\phi}_i = \frac{1}{N} \sum_{k=1}^{N} \phi_{ik} \tag{19}$$

$$\eta_i^2 = \frac{\text{SSB}_i}{\text{SST}_i} = 1 - \frac{\text{SSW}_i}{\text{SST}_i} = 1 - \frac{\sum_{g=1}^{K} \sum_{k \in \mathcal{G}_g} \left(\phi_{ik} - \bar{\phi}_{i,g}\right)^2}{\sum_{k=1}^{N} \left(\phi_{ik} - \bar{\phi}_i\right)^2} \tag{20}$$

Where SST (total sum of squares) measures the overall dispersion of SHAP values across the entire study area, SSW (within-group sum of squares) captures residual dispersion within each group, and SSB (between-group sum of squares) captures systematic differences across groups. Statistics are used to test the significance of between-group differences, in which a larger F value indicates greater inter-group divergence in feature mechanisms.

# 3. RESULTS

## 3.1 Study Setting and Analytical Sample, and GWRF Predictive Performance

The study area comprises Fangshan District and Mentougou District in Beijing, encompassing 779 communities with a total area of 3,467 km2 and population of 1.707 million, as shown in Figure 2. The two districts were among the areas most severely affected by the extreme rainstorms and floods on July 31, 2023, and exhibit substantial variation in topography, land use, infrastructure provision, and socioeconomic conditions. This heterogeneous setting provides an appropriate context for examining spatial variation in the prediction and local attribution of community resilience.

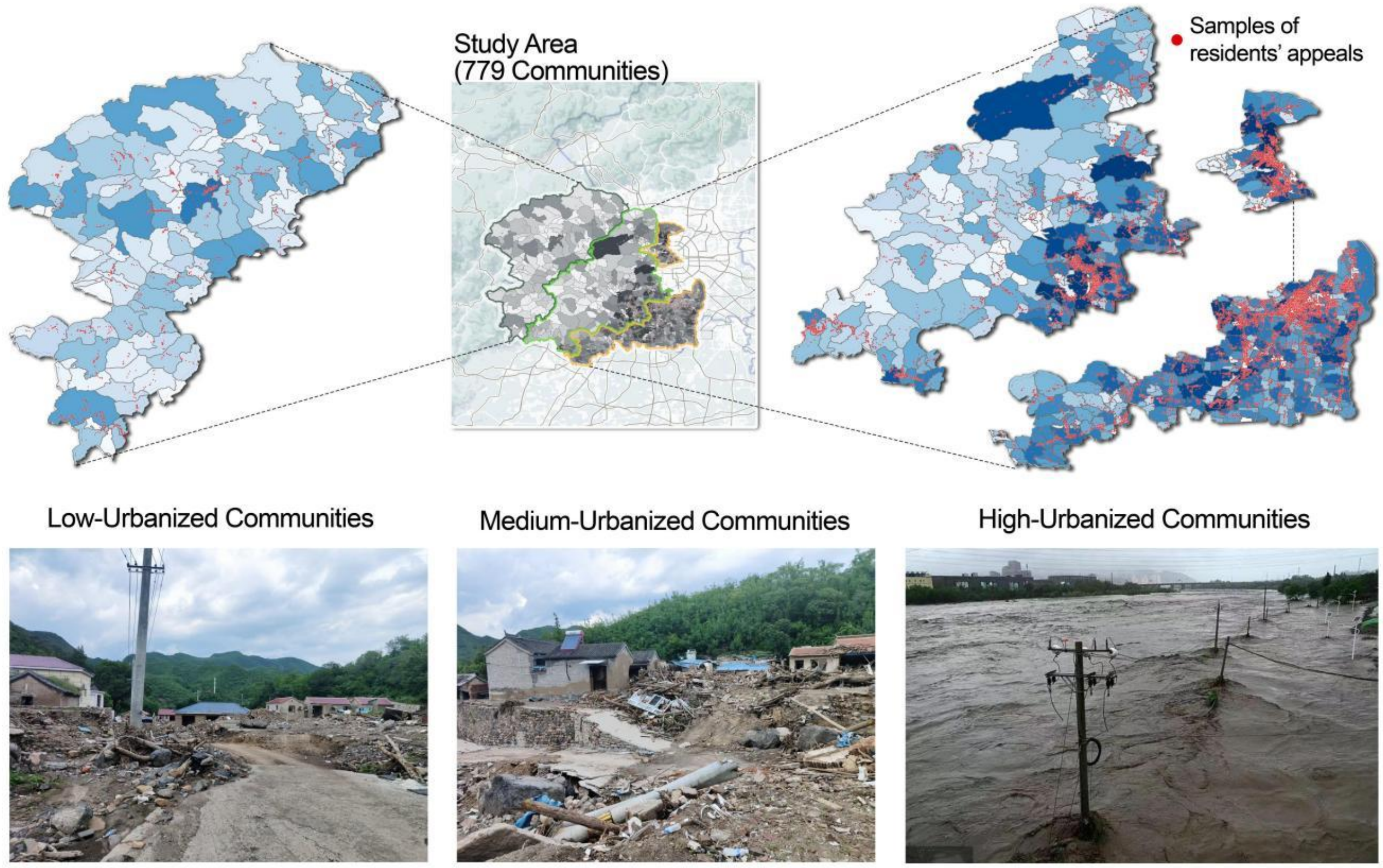

Figure 2. Overview of the study area.

Prior to modelling, all spatial datasets were harmonized to the community scale. The complete variable definitions, correlation matrix, and statistics analysis are provided in Supplementary S2.1.

The final analytical dataset comprised 779 community-level units across the two districts. The outcome variable, perceived community resilience, was obtained from resident-generated records previously and aggregated to the community level [42]. Following the data processing and index construction procedure established in our previous study [43], the resilience score was linked to the 779 community-level units used in the present analysis. The explanatory variables were organized into three dimensions: natural environment, built environment, and socioeconomic conditions. The natural environment data include precipitation, river network density, and land cover diversity, which characterize meteorological exposure and surface hydrological conditions. Built environment data include impervious surface coverage, land-use diversity, road network density, and the distribution of public service and commercial facilities, which reflect physical infrastructure and post-disaster resource accessibility. Socio-economic conditions are represented by population density, which captures the spatial distribution of residents.

The geographically weighted random forest model was used as the predictive foundation for the subsequent local attribution analysis in Local Spatial TreeSHAP, with three candidate predictors considered at each split and an adaptive neighborhood of 50 nearest-neighbor communities. The bandwidth was selected because it achieved the highest mean local validation R2 among the candidate neighborhood sizes (Supplementary Table S4). GWRF achieved a R2 of 0.741, an MAE of 1.661, and an RMSE of 2.478. Compared with the global RF benchmark, GWRF increased R2 by 0.440 and reduced MAE and RMSE by 47.7% and 36.0%. These results justify the use of GWRF as the predictive model of the attribution analysis. Comparison of benchmark models and details are reported in Supplementary Table S5.

Table 2. Hyperparameter settings of the model.

| **Parameter** | **Final value** | **Description** |
|---|---|---|
| Number of trees per local forest | 100 | Number of decision trees fitted for each focal location |
| Candidate predictors at each split | 3 | Number of randomly selected predictors considered at each tree split |
| Adaptive neighborhood bandwidth | 50 | Number of neighboring communities used in each local model |
| Maximum tree depth | 20 | Maximum depth allowed for individual decision trees |
| Random seed | 123 | Seed used to ensure reproducibility |
| Spatial weighting function | bi-square kernel | Kernel used to weight neighboring observations |

| Parameter | Final value | Description |
|---|---|---|
| Predictor number | 9 | Total number of explanatory variables used in the model |

## 3.2 Local Reference Grouping and Nonlinear Predictive Attribution Patterns

Figure 3 provides spatial distribution and nonlinear predictive associations identified by the traditional GWRF model and subsequent SHAP analysis. While partial dependence analysis captures overall nonlinear trends, it often masks true local mechanisms in spatially heterogeneous urban systems. Positive relationships observed at the global level may not hold locally and can even reverse in specific areas. This limitation arises from two key issues.

First, systematic bias introduced by the global baseline. In explainability analysis, the baseline is typically defined as the global mean prediction E[f(x)]. According to the additive decomposition principle, for high-resilience communities (where the value of dependent variable is significantly above the mean), the model tends to allocate the difference mechanically across features. Moreover, systematically inflated SHAP values can be shown (as shown by the high-value clusters in the upper part of Figure 3). This can result in misinterpretation of feature contributions, where researchers may incorrectly infer that high-resilience areas are more sensitive to all factors. This inflation primarily reflects differences in baseline development levels across regions, rather than the true local marginal effects of the variables.

Second, random noise is introduced by direct localization. To eliminate the bias caused by global baselines, a local perspective is required, where each region is compared against its own local baseline. However, directly examining ungrouped local SHAP dependence plots often yields highly scattered and disordered patterns. While localization removes systematic global bias, it introduces substantial local noise. When heterogeneous samples are mixed without a shared reference framework, it becomes difficult to extract clear and consistent patterns.

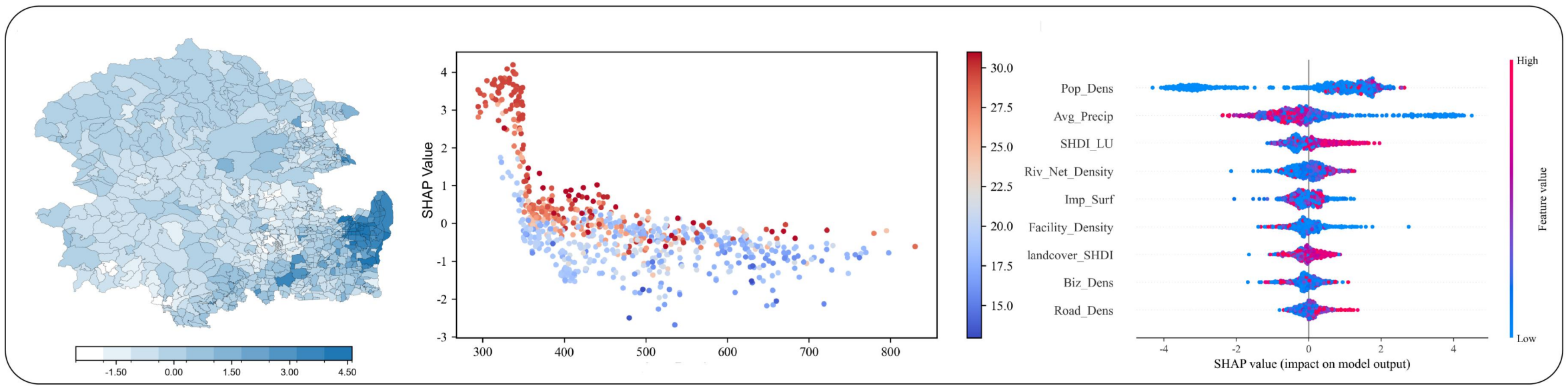

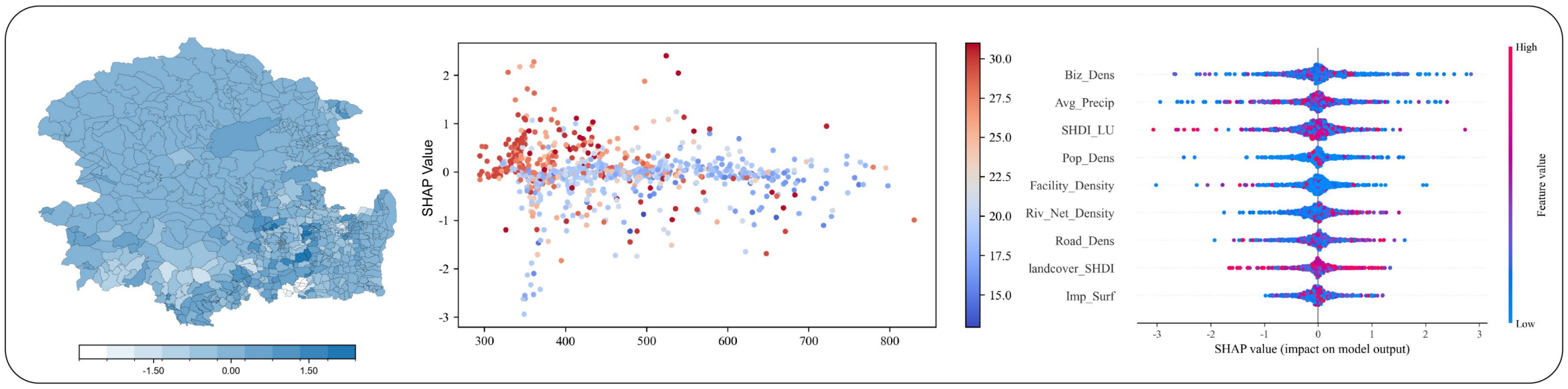

Figure 3. Spatial distribution, nonlinear marginal effects, and feature importance with traditional SHAP.

To balance global bias and local noise, this study introduces a spatial grouping strategy aimed at aggregating samples with similar underlying mechanisms. Two approaches are compared: K-means clustering based on feature values (data-driven) and stratification based on urbanization levels (knowledge-driven). Variance analysis indicates that grouping based on urbanization levels achieves better performance, with lower within-group variance in SHAP values and more pronounced between-group differences. This result is also theoretically justified, as urbanization level reflects infrastructure capacity, public service provision, and accumulated social capital, which are key contextual factors influencing post-disaster recovery mechanisms. Accordingly, the study area is divided into three groups (high, medium, and low urbanization) to uncover distinct nonlinear response patterns across different development stages.

Table 2 compares the spatial autocorrelation intensity and the number of significant spatial units between the original predictor values and their corresponding Local Spatial TreeSHAP values. The Local Spatial TreeSHAP values showed lower Moran values than the original values for all nine variables. Across nine variables, the values declined by an average of 66.2%, and the number of significant spatial units declined by an average of 65.5%. For example, the Moran values of impervious surface ratio decreased from 0.5437 for the original predictor values to 0.0927 for the Local Spatial TreeSHAP values.

Table 2. Comparison of spatial autocorrelation intensity between original variables and TreeSHAP values.

| Variable | Moran's I (original) | Moran's I (SHAP) | Decline in Moran's I | Significant Clusters (original) | Significant Clusters (SHAP) | Reduction in Significant Clusters |
|---|---|---|---|---|---|---|
| Avg_Precip | 0.5536 | 0.1296 | 76.58% | 566 | 152 | 73.14% |
| Biz_Dens | 0.5596 | 0.1516 | 72.92% | 608 | 201 | 66.94% |
| Facility_Density | 0.3286 | 0.0644 | 80.40% | 524 | 128 | 75.57% |
| Imp_Surf | 0.5437 | 0.0927 | 82.95% | 583 | 198 | 66.04% |
| Landcover_SHDI | 0.3608 | 0.1298 | 64.03% | 421 | 205 | 51.31% |
| Pop_Dens | 0.4885 | 0.2026 | 58.53% | 641 | 262 | 59.13% |
| Riv_Net_Density | 0.1368 | 0.0171 | 87.53% | 359 | 90 | 74.93% |
| Road_Dens | 0.0769 | 0.0716 | 6.92% | 283 | 134 | 52.65% |
| SHDI_LU | 0.3972 | 0.1359 | 65.79% | 551 | 166 | 69.87% |

*Note: Significant spatial units refer to units identified as statistically significant in the local spatial autocorrelation analysis at $p<0.05$*

As shown in Figures 4, the grouped TreeSHAP dependence plots reveal distinct nonlinear trajectories of the same variables across different urbanization levels, confirming that global average patterns obscure local heterogeneous mechanisms. The vertical axis represents TreeSHAP contribution values, while color intensity indicates feature magnitude. The high-urbanization group exhibits large fluctuations in SHAP values on both positive and negative sides; the medium-urbanization group shows an overall shift toward negative values with reduced amplitude; and the low-urbanization group displays SHAP values almost entirely in the negative range.

Natural environmental factors exhibit stronger sensitivity and threshold effects in low-urbanization areas. The precipitation dependence plot shows clear gradient differences across groups. In highly urbanized areas, SHAP values fluctuate around zero at low rainfall levels, indicating that engineered infrastructure partially mitigates rainfall impacts; significant negative effects emerge only when rainfall exceeds approximately 500 mm. In medium-urbanization areas, the overall curve shifts downward, with negative effects appearing earlier. In low-urbanization areas, SHAP values remain negative across the entire range and decrease monotonically with increasing rainfall. A similar pattern is observed for impervious surface ratio: while high-urbanization areas maintain relatively stable SHAP values despite high imperviousness, low-urbanization areas exhibit a sharp decline in SHAP values as impervious surfaces increase.

Among built environment factors, public service facility density shows one of the most pronounced cross-group differences. In highly urbanized areas, facility density is already high, and SHAP values fluctuate around zero, indicating that additional investment yields negligible marginal returns. In medium-urbanization areas, positive effects are present but limited. In low-urbanization areas, SHAP values are almost entirely negative with large magnitudes, reflecting severe shortages of public services rather than negative intrinsic effects.

Furthermore, land-use and land-cover indices exhibit clear cross-group differences. In highly urbanized areas, the land-use index shows a wide range of variation, with positive SHAP values concentrated in the mid-range (approximately 0.75 – 1.25), indicating the beneficial role of functional diversity. In medium-urbanization areas, the positive range narrows. In low-urbanization areas, SHAP values are entirely negative and concentrated within a narrow range, reflecting limited land-use diversity and structural constraints. The land-cover index follows a similar pattern, where moderate landscape diversity supports resilience in highly urbanized areas, but its regulatory potential is minimal in low-urbanization regions.

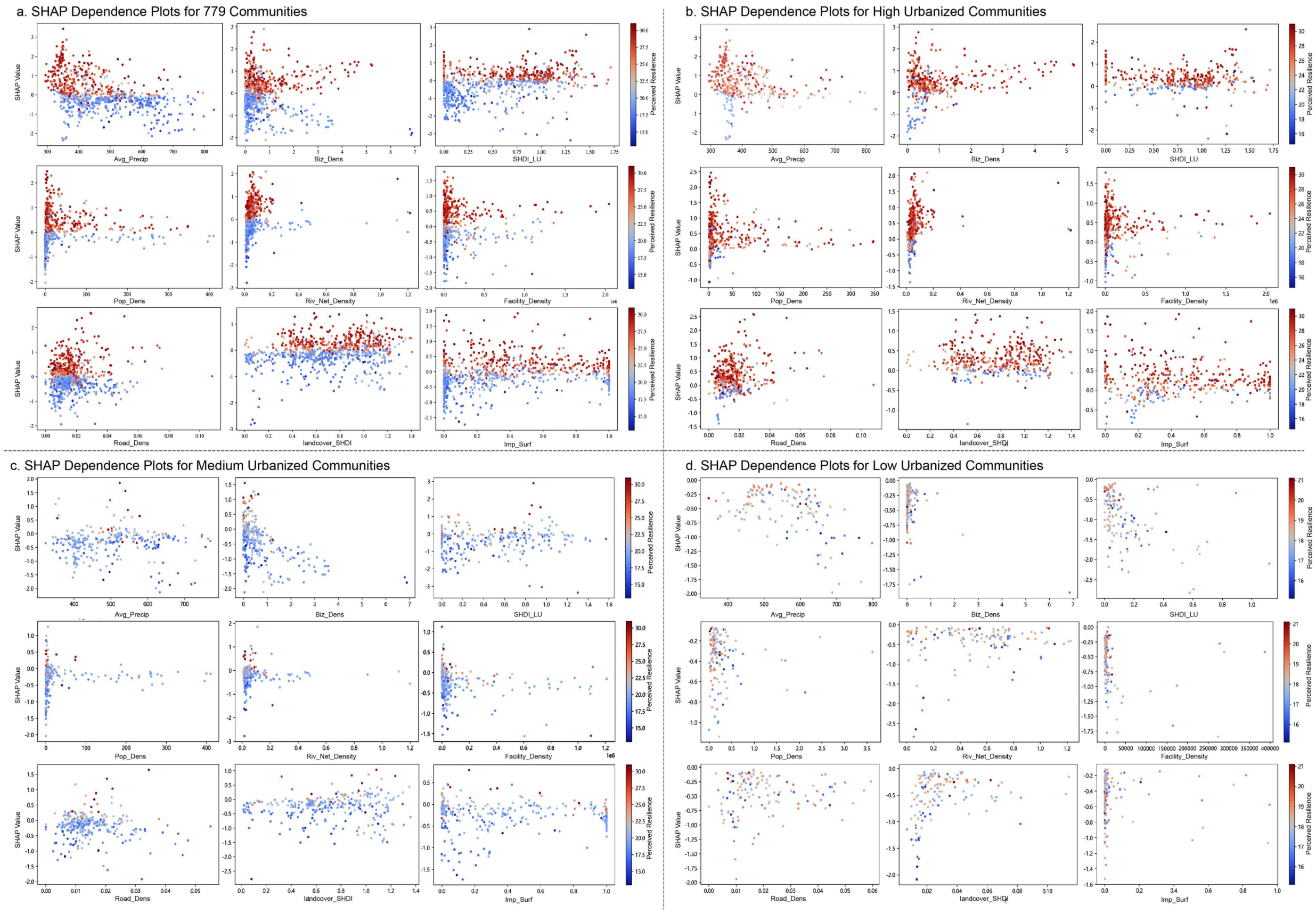

Figure 4. Feature contributions in the three urbanization groups.

### 3.3 Decomposing Multi-factor Main and Interaction Effects Based on TreeSHAP

The preceding analysis identified key influencing factors and their nonlinear marginal effects. However, urban resilience fundamentally represents a complex system of interdependent components, in which the effect of any single factor is often modulated and constrained by others. To uncover the underlying nonlinear synergistic and antagonistic mechanisms, this section applies the TreeSHAP interaction decomposition method, which separates the total SHAP effect of each feature into main effects and interaction effects.

When the main effect dominates the total contribution, the system can be characterized as single factor driven, suggesting that optimization strategies may focus on targeted interventions.

However, when interaction effects constitute a substantial proportion, the system operates under multi-factor coupling, where the impact of any individual factor depends on its coordination with others, necessitating integrated and system-level optimization strategies.

### 3.3.1 Main Effects and Interaction Effects

Figure 5-6 presents the total TreeSHAP effects of each factor and their internal composition across different urbanization groups, where the solid portions represent main effects and the striped portions denote interaction effects. A comparison across groups reveals a clear nonlinear evolution of resilience-driving mechanisms along the urbanization gradient.

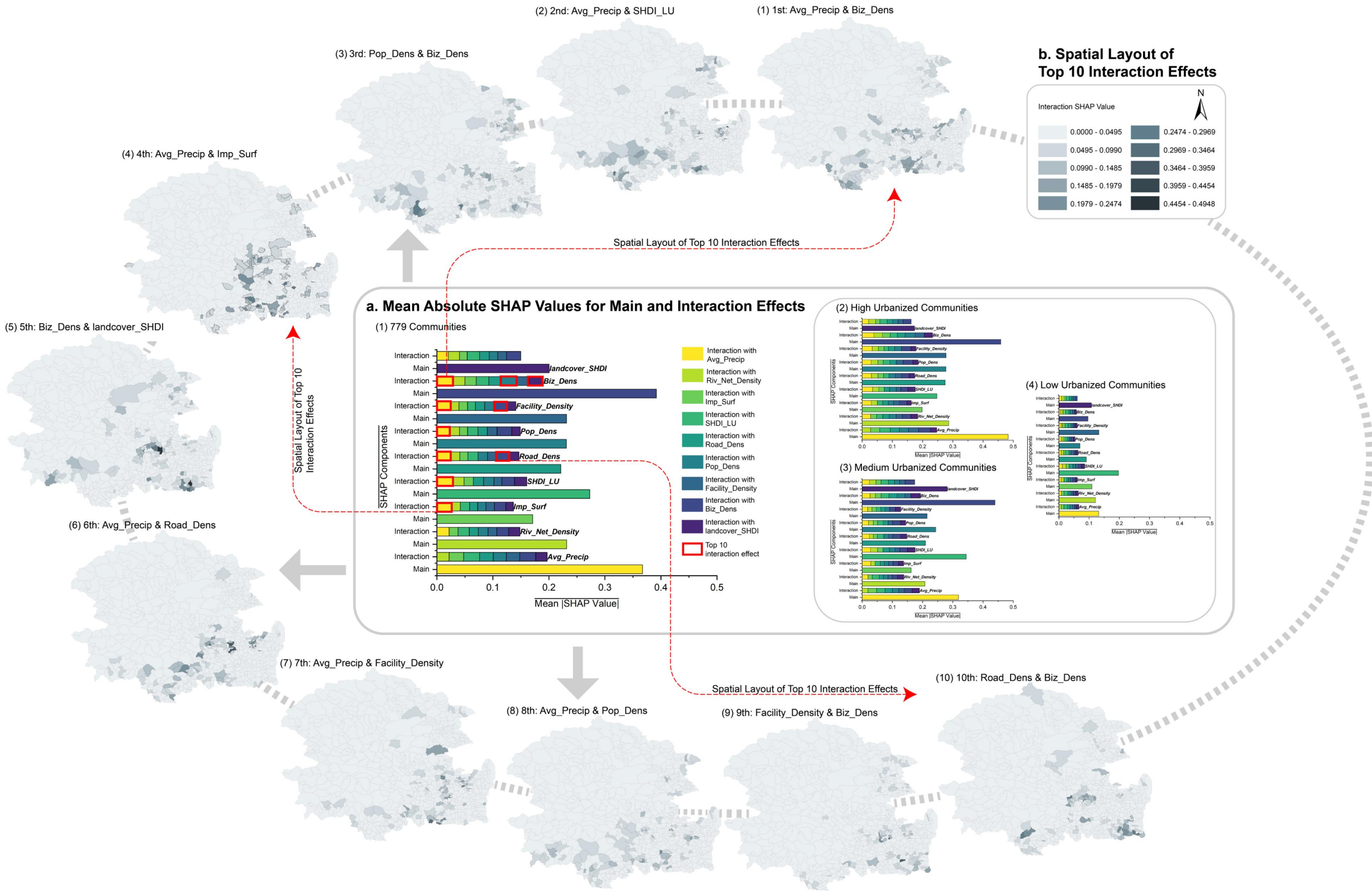

Figure 5. Main interaction effects in different spatial units. (a) Horizontal bar charts showing main effects (solid fill, representing independent contributions) and interaction effects (striped fill, representing coupled contributions). (b) Spatial distribution of dominant interaction effects (most significant interacting pairs).

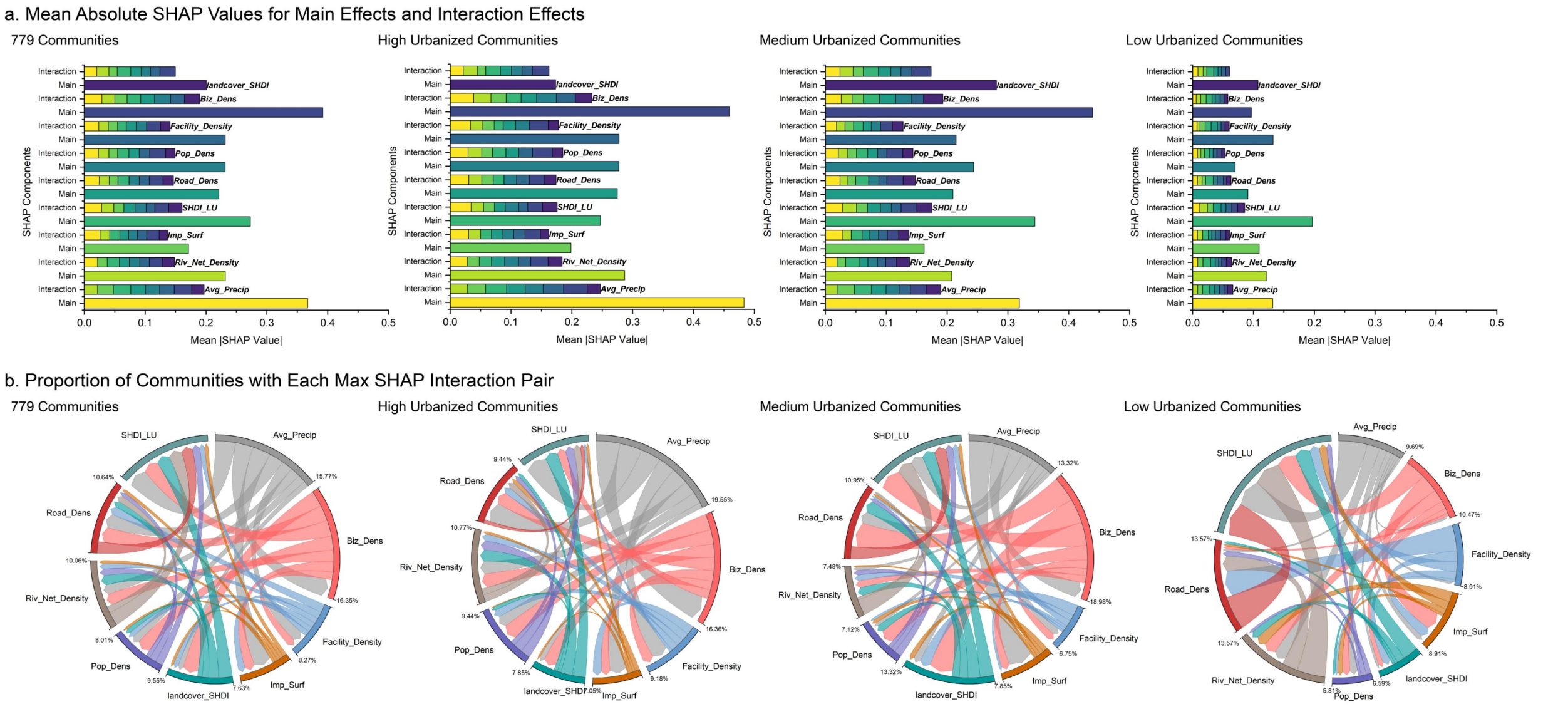

Figure 6. Main and interaction effects of influencing factors across urbanization groups. (a) Mean absolute main and interaction effects. (b) proportion of communities with each dominant interaction pair.

First, highly urbanized communities exhibit strong coupling among natural, social, and infrastructural factors. Although precipitation and commercial density rank among the most influential variables in terms of total effects, a large proportion of their contributions arises from interaction effects. This indicates that in densely built environments, factors no longer operate independently but instead rely on coordinated interactions with transportation accessibility, infrastructure provision, and socio-economic conditions. The system has evolved into a tightly interconnected co-dependent network, where the effectiveness of any single intervention depends on its alignment with surrounding conditions.

Second, medium-urbanization communities are jointly dominated by land cover and land-use factors. At this developmental stage, communities are characterized by a transitional landscape where blue–green spaces and built environments coexist. The system is no longer governed by a single dominant factor but instead depends on landscape configuration. The degree of land-use mix and the connectivity of ecological spaces emerge as critical variables for regulating flood resilience and buffering disaster impacts.

Third, low-urbanization communities are predominantly governed by natural environmental factors. Variables such as land use and river network density exhibit strong main-effect dominance, with main effects accounting for more than 60% of their total contributions. In contrast, socio-economic factors such as commercial activity and road infrastructure contribute minimally. This skewed distribution indicates a high degree of environmental sensitivity, where resilience levels are primarily constrained by natural conditions. The system lacks complex regulatory mechanisms among factors and instead displays relatively simple and monotonic relationships.

### 3.3.2 Interaction Network Structure

To further examine the underlying coordination mechanisms, Figure 7 presents chord diagrams illustrating the direction and strength of interaction effects within each urbanization group. The arc length of each variable represents its total interaction strength, while the width of connecting links indicates the magnitude of interaction between pairs of variables.

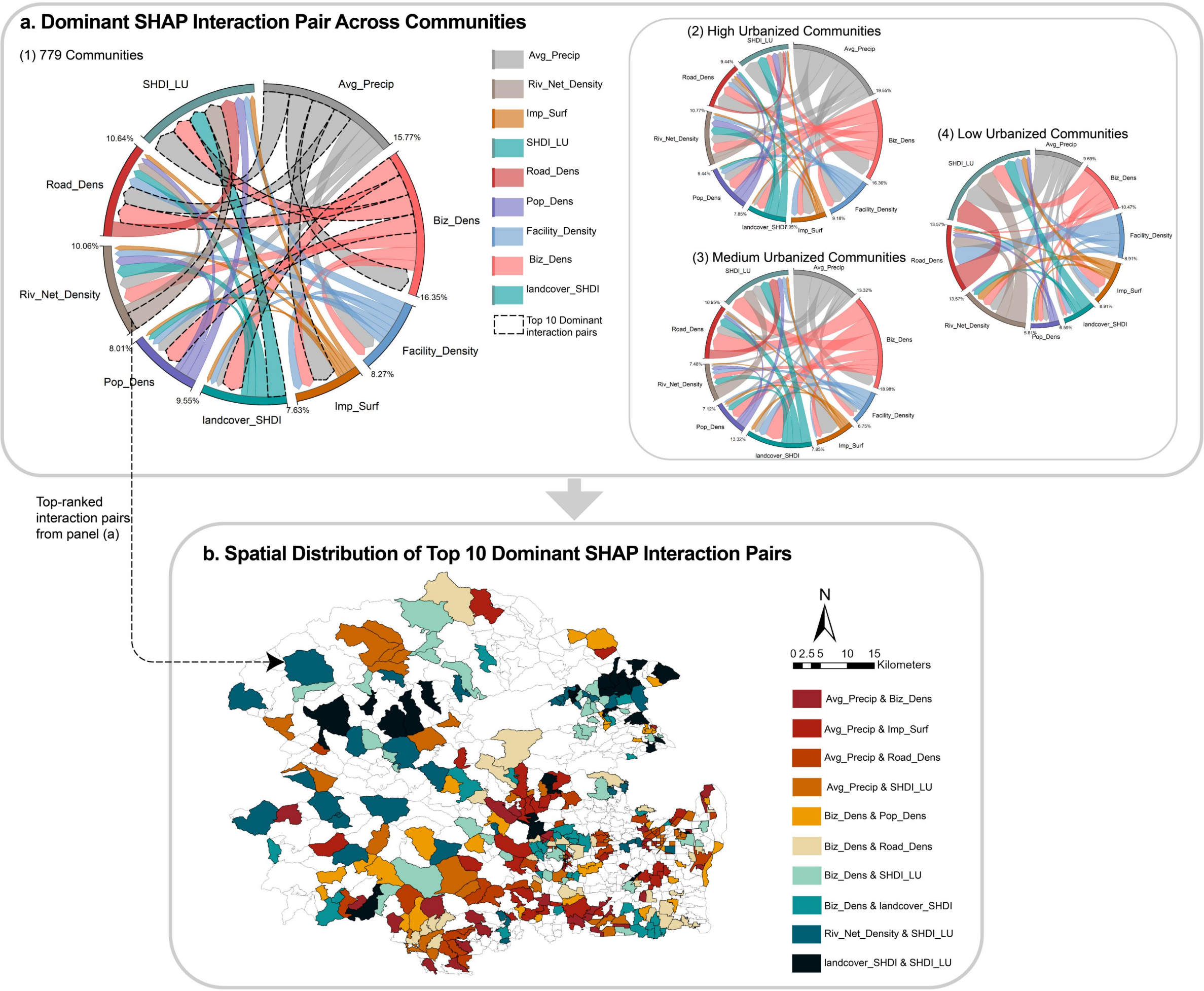

Figure 7. Interaction strength and spatial distribution of dominant interaction effects across regions. (a) Chord diagrams illustrating interaction strengths, where arc length represents total interaction intensity and link width indicates pairwise interaction magnitude. (b) Spatial distribution of dominant interaction pairs.

In high-urbanization areas, all factors are interconnected with varying strengths. The most prominent interaction occurs between road density and commercial density, indicating that the alignment between transportation accessibility and functional nodes is a key synergistic mechanism influencing resilience. Additionally, strong interactions between commercial density and population density highlight the importance of balancing service supply and demand in reducing system vulnerability.

In medium-urbanization areas, the center of the interaction network shifts toward land cover diversity. Compared to other groups, the arc length associated with land cover increases significantly and extends connections to multiple variables, including precipitation and road networks. Notably, the interaction between precipitation and land use exhibits strong intensity. This suggests that landscape configuration serves as the central coordinating mechanism, where optimizing blue–green infrastructure and land-use patterns play a crucial role in mitigating rainfall-induced risks.

In low-urbanization areas, the system exhibits a clear dominance of natural environmental interactions. The strongest interactions occur among land use, river network density, and average precipitation. In contrast, socio-economic variables such as commercial activity, infrastructure, and population density display weaker interaction strengths, as indicated by thinner arcs and links. This pattern reflects an extremely high level of environmental dependency, where the impact of rainfall is largely determined by the inherent vulnerability of the natural environment, due to the absence of mature socio-economic buffering systems.

# 4. SCENARIO ANALYSIS

## 4.1 Case-Based Analysis of Representative Communities

Three representative communities from different stages of urbanization within the study area were selected, including: (1) Community Q (Sample 52), representing highly urbanized areas; (2) Community F (Sample 368), representing medium urbanized areas; and (3) Community S (Sample 426), representing low-urbanization areas.

Community Q

2023.8.31

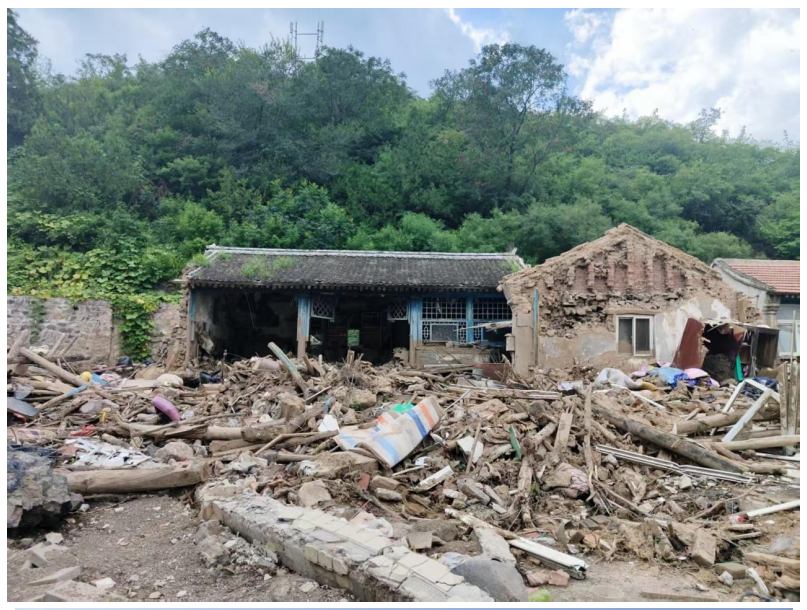

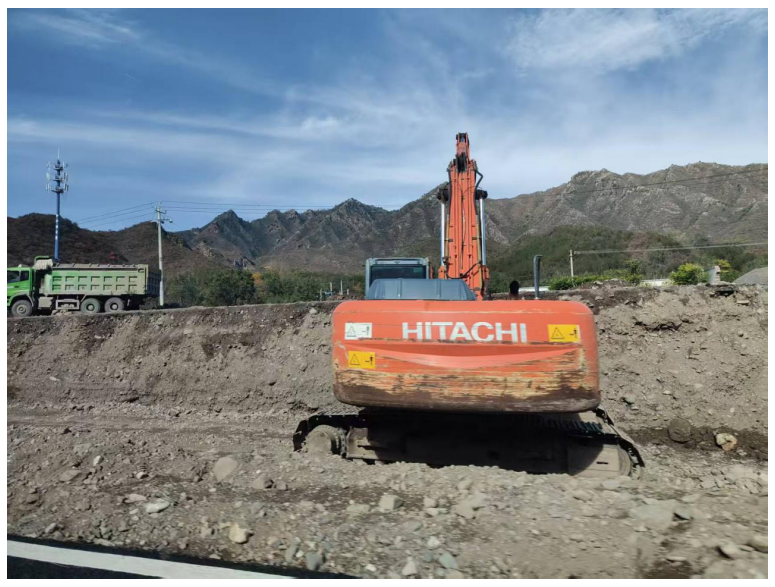

Community S

2024.4.12

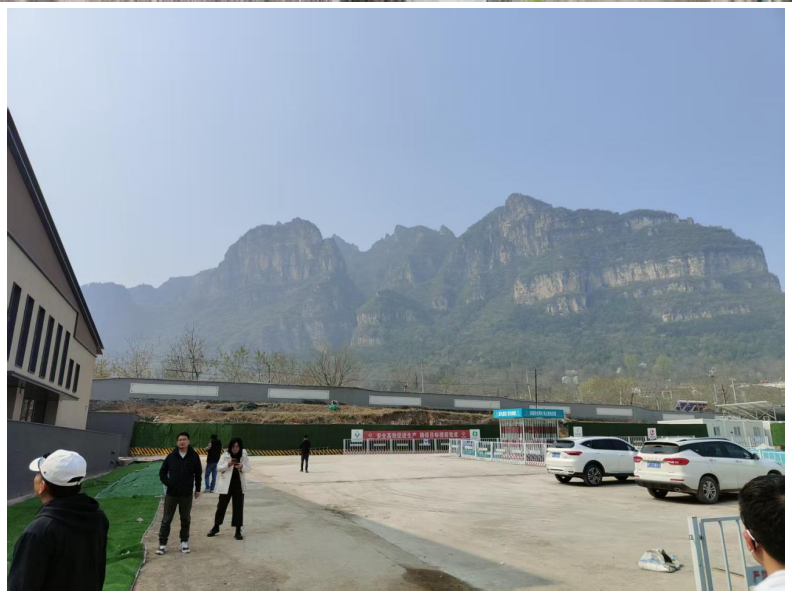

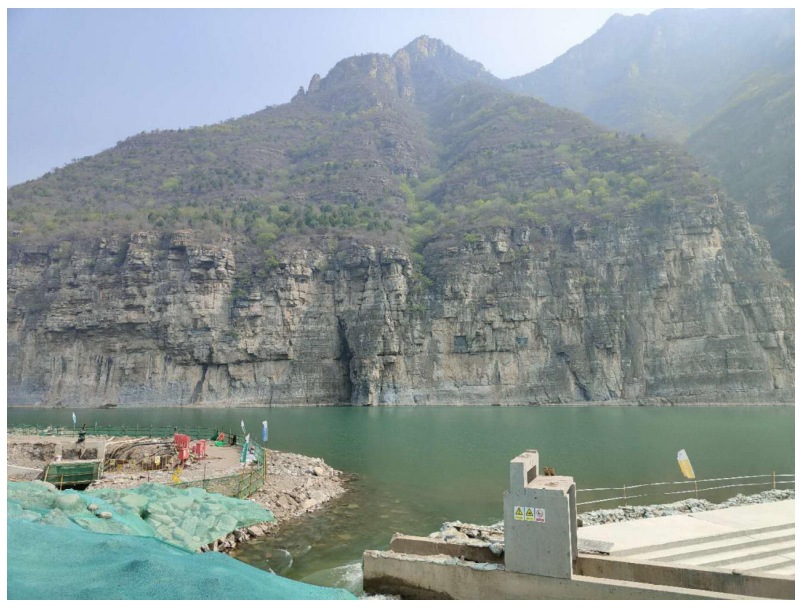

| | |
|---|---|
| Community Q | 2025.12.30 |
| Community F | 2025.12.30 |
| Community S | 2025.12.31 |

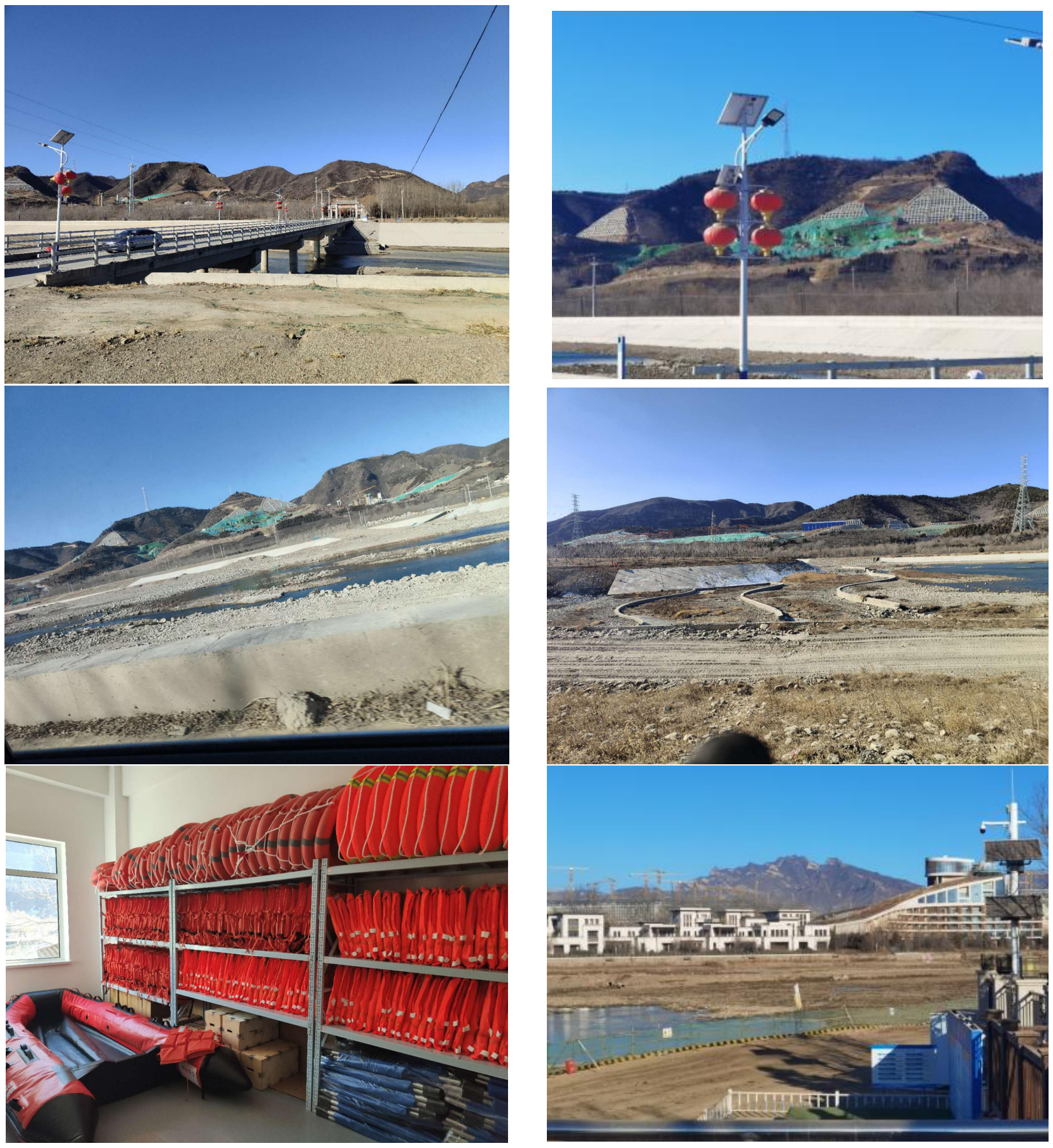

Figure 8. Field investigations across three post-disaster recovery stages in representative communities. Stage I: Immediate post-disaster period (August 2023); Stage II: Early recovery (April 2024); Stage III: Long-term reconstruction (December 2025).

Using local explanations from the TreeSHAP, including force plots, local dependence plots, and interaction heat maps (Figure 9), this section conducts micro-level attribution analyses to see how system-level coupling mechanisms manifest at the local scale.

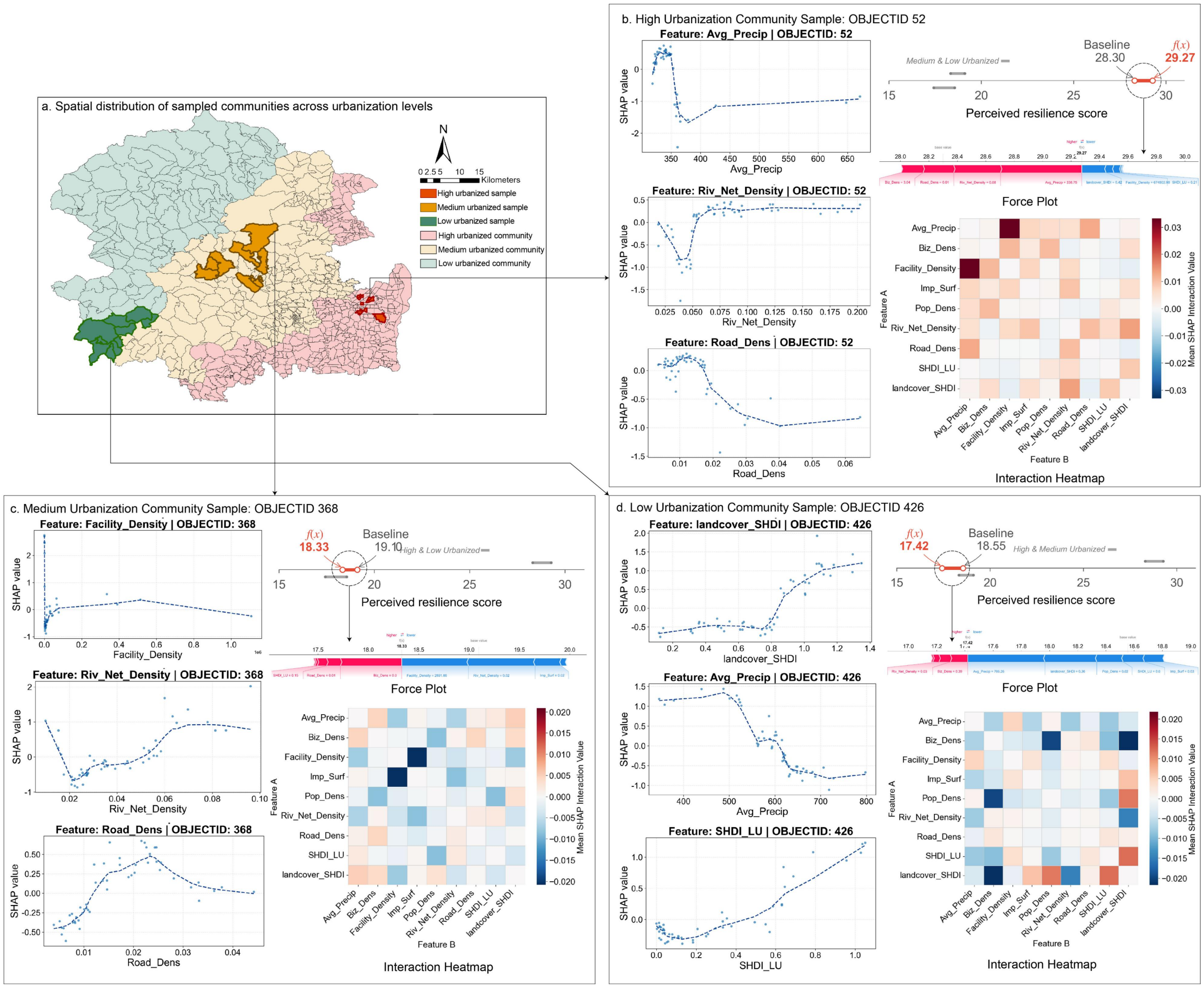

Figure 9.  Feature attribution of three representative communities based on the TreeSHAP.

### 4.1.1 High-Urbanization Case: Community Q

For Community Q, located in a highly urbanized plain area, the model predicts a resilience score of 29.27, which is substantially higher than the regional baseline, indicating strong system response capacity. According to the force plot, commercial density and river network density are the dominant positive contributors to resilience in this area. Despite high levels of population and asset exposure, the dense commercial network provides critical nodes for resource allocation, while the regulated urban water system contributes to ecological buffering and flood mitigation. The interaction heatmap further reveals a strong synergistic effect between facility density and rainfall in highly urbanized areas. This indicates that well-developed public service systems play a crucial role in emergency response and risk mitigation, enabling the system to absorb and adapt to environmental shocks.

#### 4.1.2 Medium-Urbanization Case: Community F

This community is located in a semi-mountainous zone, representing communities undergoing urbanization transformation. The predicted perceived resilience score is 18.33, lower than the regional baseline of 19.10. The primary contributors to this reduction, as indicated by the force plot, are the insufficient density of public facilities and river networks. The local dependence analysis shows that the relatively low level of facility density limits the availability of material support during disruptive events. This reflects a common issue in transitional communities, where public service provision does not keep pace with the process of urbanization.

Furthermore, interaction analysis reveals a negative antagonistic relationship between impervious surface ratio and facility density. This suggests a mismatch between development intensity and supporting infrastructure: increased surface hardening raises runoff risks, while the lack of corresponding drainage or emergency facilities exacerbates system vulnerability. As a result, development activities may unintentionally increase overall risk.

#### 4.1.3 Low-Urbanization Case: Community S

Community S, located in a deep mountainous region with low urbanization, was severely affected. TreeSHAP-based retrospective analysis shows that the predicted perceived resilience score for this community is 17.42, substantially lower than the regional baseline of 18.55. The force plot indicates that extreme rainfall intensity (700 mm) is the dominant factor driving resilience loss. The local dependence plot identifies a critical physical threshold, where marginal contributions decline sharply once rainfall exceeds approximately 550–600 mm.

Further interaction analysis reveals a significant spatial overlap between road network density and river network density. In practice, this "roads-follow-rivers" spatial configuration leads to a strong conflict between transportation corridors and flood conveyance pathways. During extreme flood events, this overlap causes a rapid breakdown of physical connectivity, transforming environmental exposure into a governance challenge where rescue operations cannot effectively reach affected areas. The negative SHAP values of the land-use SHDI in the low-value range ($<0.8$) further confirm that structural spatial monotony is a key internal driver of system collapse.

### 4.2 Field-Based Validation of Model Results

To validate the model-derived local feature importance rankings and nonlinear response patterns, qualitative verification was conducted through field interviews. Twelve semi-structured interviews were conducted with local officials, infrastructure managers, and residents. Each interview lasted approximately 30 minutes and was conducted in Mandarin. Participants were selected to cover the three urbanization settings and the functional domains relevant to post-disaster recovery. The list of interviewees is provided in Table 3.

Table 3. List of Interview Participants for Qualitative Validation

| ID | Position | Interview Focus |
|---|---|---|
| Interviewee 1 | Deputy Town Mayor | Emergency response and post-disaster recovery process |
| Interviewee 2 | Director of Reservoir Management Office | Infrastructure upgrading and engineering works |
| Interviewee 3 | Director of Drainage Affairs Center | Water supply and drainage system construction |
| Interviewee 4 | Head of Comprehensive Affairs Section | Hydraulic infrastructure recovery |
| Interviewee 5 | Local resident | Disaster experience and perceived recovery |
| Interviewee 6 | Director of Policy Research Office | Post-disaster spatial planning |
| Interviewee 7 | Deputy Township Mayor | Emergency infrastructure development |
| Interviewee 8 | Local resident | Perception of emergency facilities |
| Interviewee 9 | Section Chief, District Water Affairs Bureau | Hydraulic infrastructure construction and recovery |
| Interviewee 10 | Chair of Town People's Congress | Specific emergency facility implementation |
| Interviewee 11 | Director, District Emergency Management Bureau | Emergency facility implementation |
| Interviewee 12 | Town Mayor | Emergency support system and coordination |

#### 4.2.1 Validation in Community Q

For the community in Community Q, average precipitation was the most influential local factor (Figure 10). The local dependence plot shows a pronounced threshold effect, with SHAP values declining sharply once rainfall exceeds approximately 350–400 mm. This threshold is consistent with the flood response decision logic described by local officials (Interviewee 1).

Second, river network density ranks second, with its strongest negative effect occurring at moderate density levels. Interviewee 2 explains that the parallel structure of the Dashi and Ciwei Rivers, amplifies flood vulnerability, making river networks a critical risk factor. Commercial facility density shows a strong positive effect in the local dependence plot. Village-level commercial entities, such as small supermarkets and retail stores, can rapidly transform into decentralized supply nodes during disasters (Interviewees 1 and 3).

Finally, road network density exhibits an overall negative trend. Interviews (Interviewees 1 and 4) provide a clear explanation: although mountainous villages may have relatively dense road networks, these are typically low-grade roads with limited resilience. Many are rural or unpaved roads that are highly susceptible to disruption during disasters, and recovery requires a staged process from temporary access restoration to long-term reconstruction.

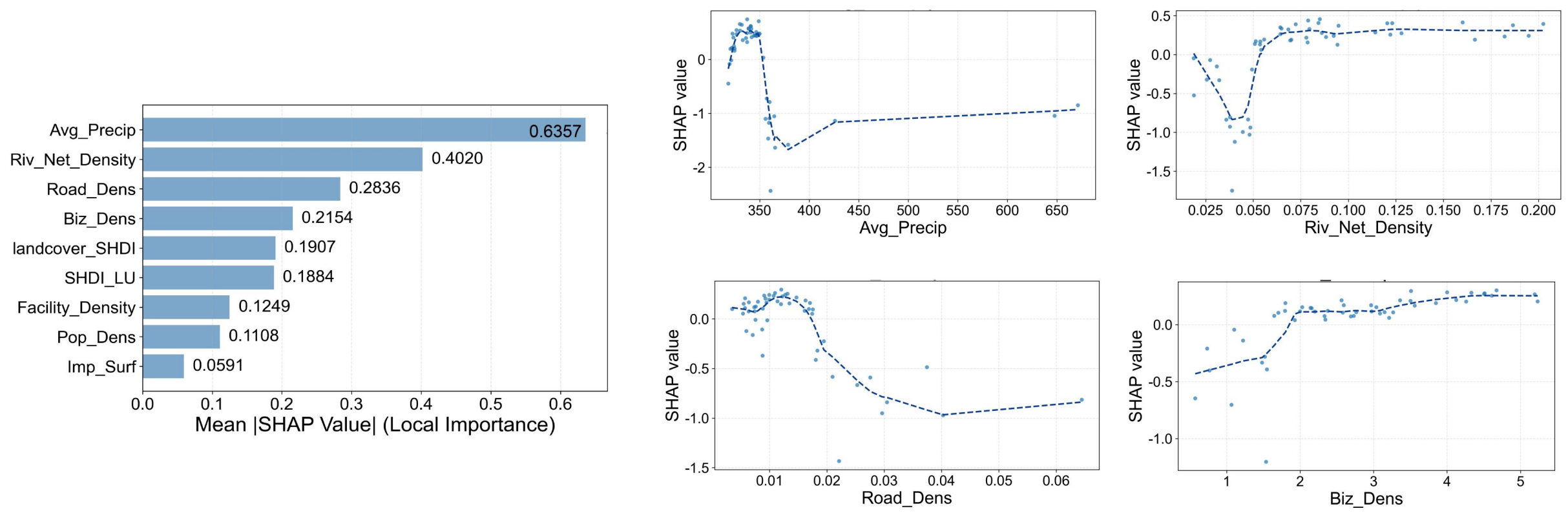

Figure 10. Local feature importance and nonlinear effects in Community Q.

### 4.2.2 Validation in Community F

In Community F, public facility density was the most important local factor, whereas precipitation ranked only eighth (Figure 11). Interviewees 5 and 6 note that communication relies on ground-based stations (which fail immediately during power outages), and water supply depends on electric pumps (which cease functioning when electricity is lost). This fragile infrastructure collapses rapidly under extreme rainfall conditions.

Second, river network density ranks second. The current density corresponds to peak flood accumulation zones without regulatory infrastructure, resulting in uncontrolled flood surges. The model's observation that resilience improves in higher-density intervals aligns with Interviewee 9's expectations regarding the potential benefits of improved flood control and storage infrastructure.

Third, road density shows an optimal range (approximately 0.01–0.03), corresponding to an ideal configuration of 2–3 redundant access routes, as described by Interviewees 7 and 8. This avoids both the vulnerability of single-route dependence and the maintenance burden associated with overly dense networks. Finally, land-use diversity exhibits a monotonic negative trend. As explained by Interviewee 7, this reflects the region's policy constraints as an ecological conservation zone, where industrial development is restricted, leading to limited land-use diversity.

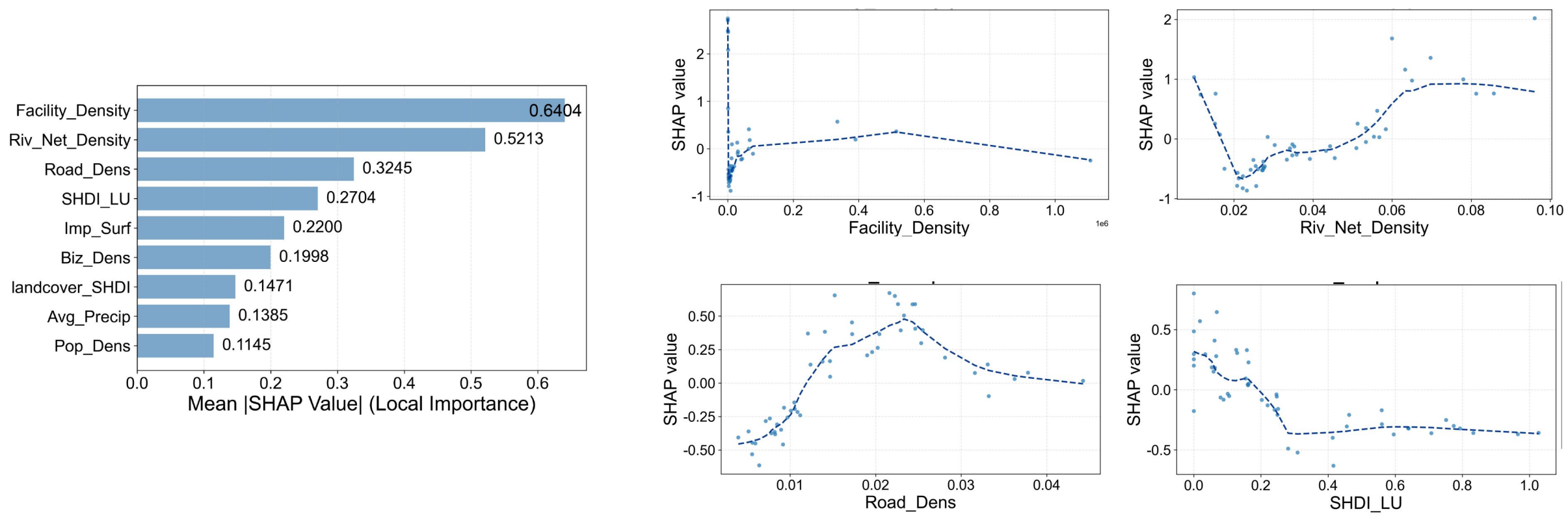

Figure 11. Local feature importance and nonlinear effects in Community F.

### 4.2.3 Validation in Community S

In Community S, SHAP results indicate that the land-cover index becomes the most important factor, making this the only case dominated by ecological variables among the three (Figure 12). The key mechanism lies in the karst terrain, where soil cannot effectively retain precipitation. As noted by Interviewees 10 and 11, under low-to-moderate vegetation diversity with exposed rock, the ecosystem lacks buffering capacity, resulting in persistently low resilience. Only when vegetation achieves sufficient density and diversity to form a multi-layered ecological buffer can effective resistance emerge.

Second, precipitation shows a different threshold pattern compared to other cases: SHAP values decline significantly only when rainfall exceeds approximately 500–600 mm, indicating a higher tolerance threshold under this specific geomorphological context.

Third, land-use diversity and commercial index rank third and fourth in importance. Interviewee 11 notes that the coexistence of homestays, tourism facilities, farmland, forests, and river systems enables flexible resource mobilization and population accommodation during disasters. However, Interviewee 12 highlights that excessive commercial density may increase relocation costs and evacuation burdens, eventually outweighing its benefits.

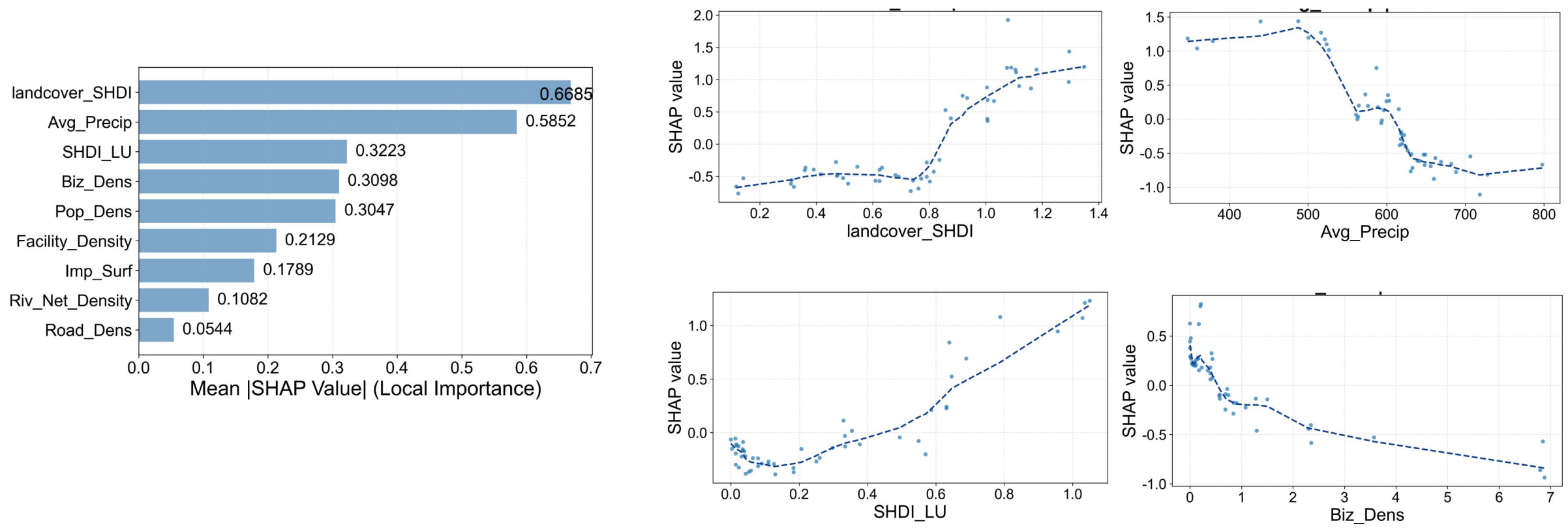

Figure 12. Local feature importance and nonlinear effects in Community S.

# 5. DISCUSSION

## 5.1 Theoretical Contributions: Resolving the Local-Global Interpretation Mismatch

This study identifies and algorithmically resolves the spatial interpretation mismatch in geographically weighted machine learning. Prior work has employed GWRF models with standard SHAP procedures without recognizing that this combination introduces a structural inconsistency: a training process that explicitly privileges local geographic context is interpreted through a background distribution that ignores that context. For instance, GWRF has been applied to crash frequency prediction [44], transport accessibility and emissions [45], street theft [46], and post-hazard public emotion [47]. When combined with additive attribution, however, existing studies typically use standard global SHAP, including applications to land finance and urban expansion [28] and human feces distribution [48]. These studies do not report the background distribution as a modelling choice, suggesting that it is treated as a default setting. Although traditional SHAP can recover spatial effects comparable to those identified by spatial lag and multiscale geographically weighted regression models [49], problems can arise when the attribution reference does not preserve the spatial weighting structure of the locally fitted model. Our Local Spatial TreeSHAP method demonstrates that correcting this inconsistency by aligning background distributions with local geographic weights throughout the attribution process can produce quantitatively and qualitatively different results that are more spatially faithful and more policy relevant.

The evidence for this claim is multifold. First, Local Spatial TreeSHAP eliminates the baseline inflation artifact observed in global SHAP applied to GWRF: high-resilience communities no longer exhibit uniformly elevated SHAP values simply because they are far above the global mean. Second, Local Spatial TreeSHAP dependence plots show sharper and more stable nonlinear structures within urbanization groups, indicating that local background distributions reduce attribution noise. Third, the spatial mapping of interaction effects derived from Local Spatial TreeSHAP reveals geographically coherent interaction regimes that accord with the known physical geography and urban development history of the study area.

In addition, urbanization gradient is validated as a principle for organizing spatially heterogeneous SHAP outputs. Such organization is necessary because location-specific attribution maps alone are difficult to interpret, and prior studies have therefore grouped locations into spatial contexts to reveal systematic shifts in predictor importance [50].The comparison against K-means clustering demonstrates that theory-informed grouping (drawing on decades of urban ecology and resilience scholarship linking urbanization stage to infrastructure dependency and natural factor sensitivity) outperforms purely data-driven grouping for mechanistic interpretation. This implies that leveraging urbanization gradient, land use transition stage, or morphological typology as conceptually grounded stratification frameworks may be better choices than blindly pursuing black-box clustering in urban spatial machine learning studies.

## 5.2 Policy Implications for Differentiated Land Use Planning

Three mechanistically distinct resilience regimes associated with urbanization levels are identified, which has direct implications for land use policy and disaster recovery governance. Prior interpretable flood studies have rarely aligned local attribution with policy [51], while differentiated regimes have mainly been identified across cities rather than within districts where policy is implemented [52].

For high-urbanization communities, the policy priority is not additional infrastructure volume but infrastructure coordination quality. SHAP saturation curves indicate that road density and commercial facility density have passed their marginal effectiveness thresholds: communities in this zone are already well-supplied with individual infrastructure components, but face vulnerabilities arising from supply-demand mismatches (e.g., commercial concentration in low-accessibility zones), spatial congestion externalities, and functional redundancy gaps in emergency scenarios.Grey infrastructure shows declining marginal effectiveness under increasingly extreme rainfall [53], while diminishing returns similarly support context-sensitive rather than uniform stormwater design standards [54]. Land use policy recommendations for this zone include: (1) requiring mixed-use development provisions in new residential approvals to maintain commercial service accessibility within 300m walking distance, aligned with the SHAP-identified saturation threshold; (2) mandating emergency facility co-location standards in commercial development permits, targeting the Facility Density × Avg_Precip interaction effect identified as the dominant positive synergy in this zone; and (3) updating stormwater management overlays to account for impervious surface accumulation that has exceeded the zone's original drainage design capacity.

For medium-urbanization communities (the transitional mountain-fringe belt), the policy priority is synchronizing development intensity with service provision, addressing the facility deficit that the interaction analysis identifies as the primary source of vulnerability. These communities are experiencing active land use conversion from rural to suburban uses, with impervious surface expansion proceeding ahead of public service infrastructure deployment. Land use policy recommendations include: (1) implementing development sequencing requirements that condition impervious surface expansion permits on concurrent provision of flood mitigation facilities (drainage, detention basins, emergency access roads), directly targeting the negative Imp Surf × Facility Density interaction identified for C Village; (2) prioritizing blue-green infrastructure investment in communities where SHAP analysis identifies land cover diversity as the dominant unmet resilience driver, including riparian buffer zone designation and green wedge protection in local urban plans; and (3) establishing sub-district resilience allocation formulas that direct post-disaster recovery funds to facility gaps rather than uniform per-capita distribution.

For low-urbanization communities (the deep mountain settlements), the policy priority is ecological threshold management, recognizing that these communities face hard natural constraints that cannot be engineered away within realistic resource envelopes, and that planning strategy must therefore focus on reducing exposure and improving connectivity redundancy. Land use policy recommendations include: (1) formally designating precipitation threshold zones (> 600 mm recurrence boundary, as identified by the Local Spatial TreeSHAP precipitation SHAP inflection point) as "elevated evacuation planning areas" in the district's risk-sensitive land use plan, triggering mandatory secondary access road provision for any settlement above this threshold with fewer than two independent road connections; (2) prohibiting new permanent settlement within the identified Road_Dens × Riv_Net_Density high-conflict zones (where road infrastructure co-located with river channels creates total connectivity loss risk), and establishing relocation incentive programs for existing residents; and (3) incentivizing land cover diversification in communities where Landcover_SHDI is below 0.8, the SHAP-identified threshold below which land cover homogeneity becomes a significant negative contributor, through ecological restoration subsidies and natural forest management programs.

At the district governance scale, the broader implication is that per-community resilience budgets and recovery allocation formulas should be recalibrated using community-specific SHAP driver profiles rather than simple exposure or damage metrics. Communities with identified threshold vulnerabilities (Group Low) should receive front-loaded pre-disaster investment in connectivity redundancy and ecological buffering; communities with facility gaps (Group Medium) should be prioritized for capital facility investment; communities in the saturation zone (Group High) should receive governance and coordination quality improvements rather than additional infrastructure volume.

## 5.3 Limitations and Future Research Directions

Although important insights are provided, several limitations of the present study merit explicit acknowledgment. First, the analysis is cross-sectional, capturing resilience at a single 30-day post-event snapshot. Resilience perceptions are dynamic (they evolve as communities observe the pace of recovery over weeks and months), and a longitudinal extension of the GWRF-SHAP framework to repeated surveys would substantially enrich understanding of recovery trajectory heterogeneity. Second, the resilience outcome variable, while theoretically validated and survey-derived, is subject to reporting biases including differential response rates across urbanization zones and social desirability effects in face-to-face administration contexts. Third, the local Spatial TreeSHAP method, while theoretically grounded and empirically validated, has been developed and tested within a single case study. Systematic evaluation across multiple geographic contexts, hazard types, and GWRF parameterizations would strengthen confidence in its generalizability and reveal any boundary conditions on its applicability. Fourth, the present analysis treats urbanization level as a fixed stratification criterion; future work could develop adaptive, data-driven stratification methods that update community typologies as development trajectories evolve.

Beyond these considerations, the framework developed here, including GWRF prediction, Local Spatial TreeSHAP attribution, urbanization-stratified interpretation, and policy translation, constitutes a replicable methodological template applicable to other disaster-prone cities in the Global South and rapidly urbanizing Asia-Pacific region. Cities characterized by similar within-city urbanization gradients (Jakarta, Dhaka, Hanoi, Lagos) face analytically equivalent challenges of spatially heterogeneous resilience drivers and spatially uniform policy frameworks. Applying Local Spatial TreeSHAP in these contexts would allow direct cross-city comparison of resilience regime typologies and interaction structures, contributing to a cumulative, generalizable science of spatially adaptive disaster governance.

## 6. CONCLUSION

This study has demonstrated that community-level resilience following an extreme rainfall event is governed by fundamentally different mechanisms depending on a community's urbanization stage, and that identifying and mapping these differences requires a methodological framework that is simultaneously locally calibrated, nonlinearly flexible, and interpretively faithful. Using 779 communities in Beijing as an empirical platform, we have shown that: (1) SHAP value spatial autocorrelation is dramatically lower than physical variable spatial autocorrelation, confirming that resilience mechanisms are spatially non-stationary even where physical endowments are spatially clustered; (2) Local Spatial TreeSHAP, by replacing global background distributions with locally normalized geographic kernel weights in tree path integration, produces attributions that are more spatially coherent and mechanistically informative than global SHAP applied to GWRF; (3) three distinct resilience regimes (synergistic coupling in high-urbanization zones, facility-gap constraint in medium-urbanization transitional communities, and ecological threshold dominance in low-urbanization mountain settlements) exhibit qualitatively different nonlinear response structures that call for qualitatively different policy responses.

The policy implications of these findings challenge the foundational assumption of spatial uniformity that underlies most disaster recovery frameworks. Our findings provide empirically grounded, community-specific guidance for post-disaster recovery investment prioritization, infrastructure sequencing, and land use regulation, including specific precipitation thresholds for elevated evacuation planning designation, facility-development sequencing requirements to address the impervious surface–facility density antagonism in transitional zones, and road rerouting priorities to reduce the catastrophic conflict in mountain villages.

More broadly, this work establishes a generalizable methodological framework that can be transferred to other rapidly urbanizing regions facing intensifying extreme weather. The framework's capacity to generate community-type-specific planning thresholds and interaction diagnostics, derived directly from survey-linked spatial machine learning rather than expert judgment, represents a step toward evidence-grounded, place-sensitive resilience governance that can meet the governance demands of an urbanizing world under changing climate.

# Local Spatial TreeSHAP: Geographically Explainable AI Revealing Urbanization-Dependent Spatial Heterogeneity in Flood Resilience

## Authors

Zeyu Zhao[a], Xiaoshan Zhou[b], Fanxin Meng[c], Dongping Fang[d,*], Helen X.H. Bao[a,*]

## Affiliations

[a] *Department of Land Economy, University of Cambridge, Cambridge, United Kingdom*

[b] *School of Project Management, University of Sydney, NSW, Australia, 2037*

[c] *State Key Laboratory of Regional Environment and Sustainability, School of Environment, Beijing Normal University, Beijing 100875, China.*

[d] *Department of Construction Management, Tsinghua University, Beijing, China.*

*Corresponding author.
Email address: fangdp@tsinghua.edu.cn; Hxb20@cam.ac.uk

## E-mail addresses

Zeyu Zhao, Email: zz521@cam.ac.uk

Xiaoshan Zhou, Email: xiaoshan.zhou@sydney.edu.au

Fanxin Meng, Email: fanxin.meng@bnu.edu.cn

Dongping Fang, Email: fangdp@tsinghua.edu.cn

Helen X.H. Bao, Email: Hxb20@cam.ac.uk

## Declaration of conflicting interest:

The authors declared no potential conflicts of interest with respect to the research, authorship, and/or publication of this article.

Supplementary materials for

# Local Spatial TreeSHAP: Geographically Explainable AI Revealing Urbanization-Dependent Spatial Heterogeneity in Flood Resilience

The Supplementary Material contains four tables and three figures.

## 1 Symbols, notation, and acronyms

The main symbols, notation, and acronyms are shown in Table S.1and Table S.2.

**Table S.1. Mathematical symbols and notation.**

| Symbol | Description |
|---|---|
| $i$, $u$ | Index of a target location, i.e. a community-level spatial unit |
| $(u_i, v_i)$ | Geographic coordinates of location $i$ |
| $s_k$ | Geographic coordinates of training sample $k$ |
| $N$ | Number of training samples ($N = 779$) |
| $M$ | Number of explanatory variables ($M = 9$) |
| $x$ | Feature vector of a community |
| $y_k$ | Observed perceived resilience of sample $k$ |
| $f_u(\cdot)$ | Localized GWRF prediction function fitted at location $u$ |
| $d_{ij}$ | Euclidean distance between observations $i$ and $j$ |
| $b$ | Kernel bandwidth of the adaptive bi-square kernel |
| $w_{ij}$ | Geographic weight assigned to observation $j$ when fitting the local model at $i$ |
| $w_k(u)$ | Normalized geographic weight of sample $k$ at location $u$, with $\Sigma_k w_k(u) = 1$ |
| $\pi_u$ | Locally weighted background (reference) distribution induced at location $u$ |
| $E_{\pi_u}[\cdot]$ | Expectation taken under the local distribution $\pi_u$ |
| $S$ | Subset (coalition) of features excluding the feature under evaluation |
| $\varphi_0$ | Global baseline, i.e. expected model output over the whole training set |
| $\varphi_0(u)$ | Local baseline at location $u$, defined as $E_{\pi_u}[y]$ |
| $\varphi_j$ | SHAP value (attribution) of feature $j$ |
| $\varphi_j^{(u)}(x)$ | Local Spatial TreeSHAP value of feature $j$ for sample $x$ at location $u$ |
| $v_u(S)$ | Coalition value function evaluated under $\pi_u$ |
| $n$ | A node of a decision tree (root, internal or leaf) |
| $C_u(n)$ | Geographically weighted coverage of node $n$, i.e. the sum of geographic weights of training samples reaching $n$ |
| $T$ | Number of trees in the local ensemble |
| $L$ | Maximum number of leaves per tree |
| $D$ | Maximum tree depth |
| $\Phi^{(u)}$ | $M \times M$ symmetric SHAP interaction matrix at location $u$ |
| $\varphi_{ii}$ | Main effect of feature $i$, i.e. the diagonal element of $\Phi^{(u)}$ |
| $\varphi_{ij}$ | Pairwise interaction effect between features $i$ and $j$ |
| $I$ | Global Moran's $I$, a measure of spatial autocorrelation |
| SST, SSW, SSB | Total, within-group and between-group sums of squares of SHAP values |
| $\eta^2$ | Proportion of SHAP-value variance explained by a grouping scheme, $\eta^2$ = SSB / SST |
| $F$ | $F$-statistic testing the significance of between-group differences |
| $R^2$ | Coefficient of determination under spatial cross-validation |

**Table S.2. Acronyms.**

| Acronym | Definition |
|---|---|
| AI | Artificial intelligence |
| CV | Cross-validation |
| GWR | Geographically weighted regression |
| GWRF | Geographically weighted random forest |
| LISA | Local indicators of spatial association |
| MAE | Mean absolute error |
| MGWR | Multiscale geographically weighted regression |
| OLS | Ordinary least squares |
| PDP | Partial dependence plot |
| RF | Random forest |
| RMSE | Root means square error |
| SHAP | Shapley additive explanations |
| SHDI | Shannon diversity index |
| TreeSHAP | Tree-structured exact algorithm for computing SHAP values |
| VIF | Variance inflation factor |
| XAI | Explainable artificial intelligence |

## 2 Data preprocessing and results

All explanatory variables were aligned to the 779 community-level analytical units in Fangshan and Mentougou Districts. Spatial datasets were projected to a common coordinate reference system and clipped to the study-area boundary. Predictor values were then derived or aggregated for each community polygon.

For raster-based variables, including average precipitation and impervious surface ratio, values were calculated using zonal statistics. For line-based variables, including river-network density and road density, the total length of the relevant network within each community was divided by community area. For point-based variables, including public-service facilities and commercial facilities, densities were calculated as the number of facilities per unit area. Population density was calculated as the resident population divided by community area. All derived variables were checked for missing values, spatial mismatches, and extreme values before model fitting. The full definitions, units, temporal references, and data sources are provided in Table S.3.

**Table S.3. Summary of explanatory variables**

| Category | Variable | Definition | Data Source | Unit | Ref |
|---|---|---|---|---|---|
| Precipitation | Avg_Precip | Average precipitation | China Meteorological Administration (https://bj.cma.gov.cn/) | mm | [1] |
| River Network | Riv_Net_Density | River network density | NASA (https://www.nasa.gov/) | km/km² | [2,3] |
| Land Cover | landcover_SHDI | Landscape diversity index | Wuhan University SinoLC-1 (https://doi.org/10.5281/zenodo.707461) | Index | [4] |
| Impervious Surface | Imp_Surf | Percentage of impervious surface area | Wuhan University CLCD (http://doi.org/10.5281/zenodo.4417809) | % | [5] |
| Land Use | SHDI_LU | Land use diversity index | Tsinghua University EULUC-China (https://data-starcloud.pcl.ac.cn/earthdata/) | Index | [6] |
| Road Network | Road_Dens | Road density | OpenStreetMap (https://www.openstreetmap.org/) | km/km² | [7] |
| Population | Pop_Dens | Population density | LandScan (https://landscan.ornl.gov/) | Person/km² | [8] |
| Public Facilities | Facility_Density | Density of public service facilities | Amap API (https://lbs.amap.com/) | No./km² | [9] |
| Commercial Facilities | Biz_Dens | Density of commercial and entertainment facilities | Amap API (https://lbs.amap.com/) | No./km² | [10,11] |

Figure S1 summarizes the distributions of the explanatory variables. Most built-environment indicators, including river network density, road density, population density, facility density, and business density, are positively skewed, indicating substantial spatial heterogeneity across communities. Precipitation and land-cover diversity show relatively broader distributions, while impervious surface and land-use diversity exhibit more heterogeneous patterns.

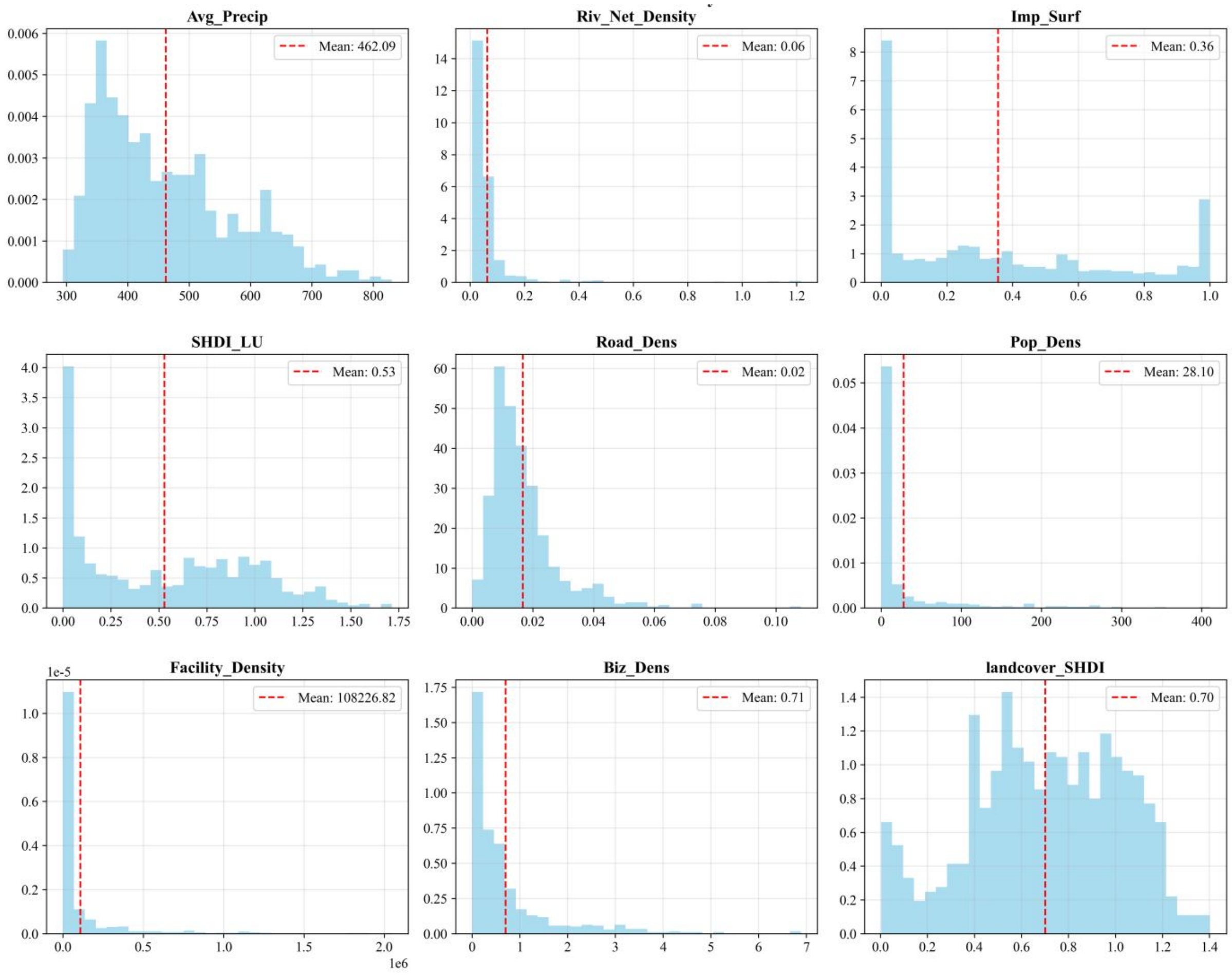

Figure S1. Distributions of the explanatory variables used in the analysis.

To assess potential multicollinearity among the retained explanatory variables, Pearson correlation coefficients were calculated. An absolute correlation coefficient of 0.70 was used as the threshold for strong pairwise correlation. All coefficients remained below this threshold, with the maximum absolute correlation of 0.642, indicating no severe pairwise collinearity among the selected variables.

**Table S.4. Pearson correlation coefficients among explanatory variables**

| | Avg_Precip | Riv_Net_Density | Imp_Surf | SHDI_LU | Road_Dens | Pop_Dens | Facility_Density | Biz_Dens | landcover_SHDI |
|---|---|---|---|---|---|---|---|---|---|
| Avg_Precip | 1.000 | | | | | | | | |
| Riv_Net_Density | -0.165 | 1.000 | | | | | | | |
| Imp_Surf | -0.642 | 0.407 | 1.000 | | | | | | |

| | Avg_Precip | Riv_Net_Density | Imp_Surf | SHDI_LU | Road_Dens | Pop_Dens | Facility_Density | Biz_Dens | landcover_SHDI |
|---|---|---|---|---|---|---|---|---|---|
| SHDI_LU | -0.333 | -0.086 | 0.106 | 1.000 | | | | | |
| Road_Dens | -0.010 | -0.061 | -0.102 | -0.049 | 1.000 | | | | |
| Pop_Dens | -0.253 | 0.279 | 0.524 | -0.174 | -0.090 | 1.000 | | | |
| Facility_Density | -0.340 | 0.221 | 0.621 | -0.121 | -0.081 | 0.603 | 1.000 | | |
| Biz_Dens | -0.260 | 0.189 | 0.445 | 0.010 | -0.098 | 0.620 | 0.569 | 1.000 | |
| landcover_SHDI | -0.181 | -0.004 | 0.048 | 0.461 | -0.093 | -0.146 | -0.136 | -0.051 | 1.000 |

Model validation was designed to prevent information leakage between model fitting, bandwidth selection, and performance assessment. The 779 communities were partitioned into ten validation folds using stratified spatial partitioning. For each fold, models were fitted using only the training communities, while predictions were generated for the held-out communities. The focal validation community was not included in its own local training neighbourhood. Candidate adaptive bandwidths from 20 to 120 were evaluated using the training-validation procedure, and a bandwidth of 50 was retained because it provided the best balance between overall and local predictive performance. All benchmark models, including OLS, GWR, RF, and GWRF, were evaluated using the same partitions and response variable.

The predictive performance of GWRF was further compared with OLS, GWR, and RF as benchmark models (Table S.5). GWRF substantially outperformed all benchmarks, achieving the highest $R^2$ and the lowest MAE and RMSE.

**Table S.5. Performance comparison with benchmark models.**

| Model | $R^2$ | MAE | RMSE |
|---|---|---|---|
| OLS | 0.344 | 3.275 | 3.994 |
| GWR | 0.274 | 3.231 | 4.113 |
| RF | 0.301 | 3.176 | 3.870 |
| GWRF | **0.741** | **1.661** | **2.478** |

## 3 Model Performance

To further examine the nonlinear relationships captured by GWRF, partial dependence plots were generated for the retained explanatory variables (Figure S2). The results reveal clear nonlinear, threshold, and saturation patterns across several predictors, supporting the use of a nonlinear modelling framework.

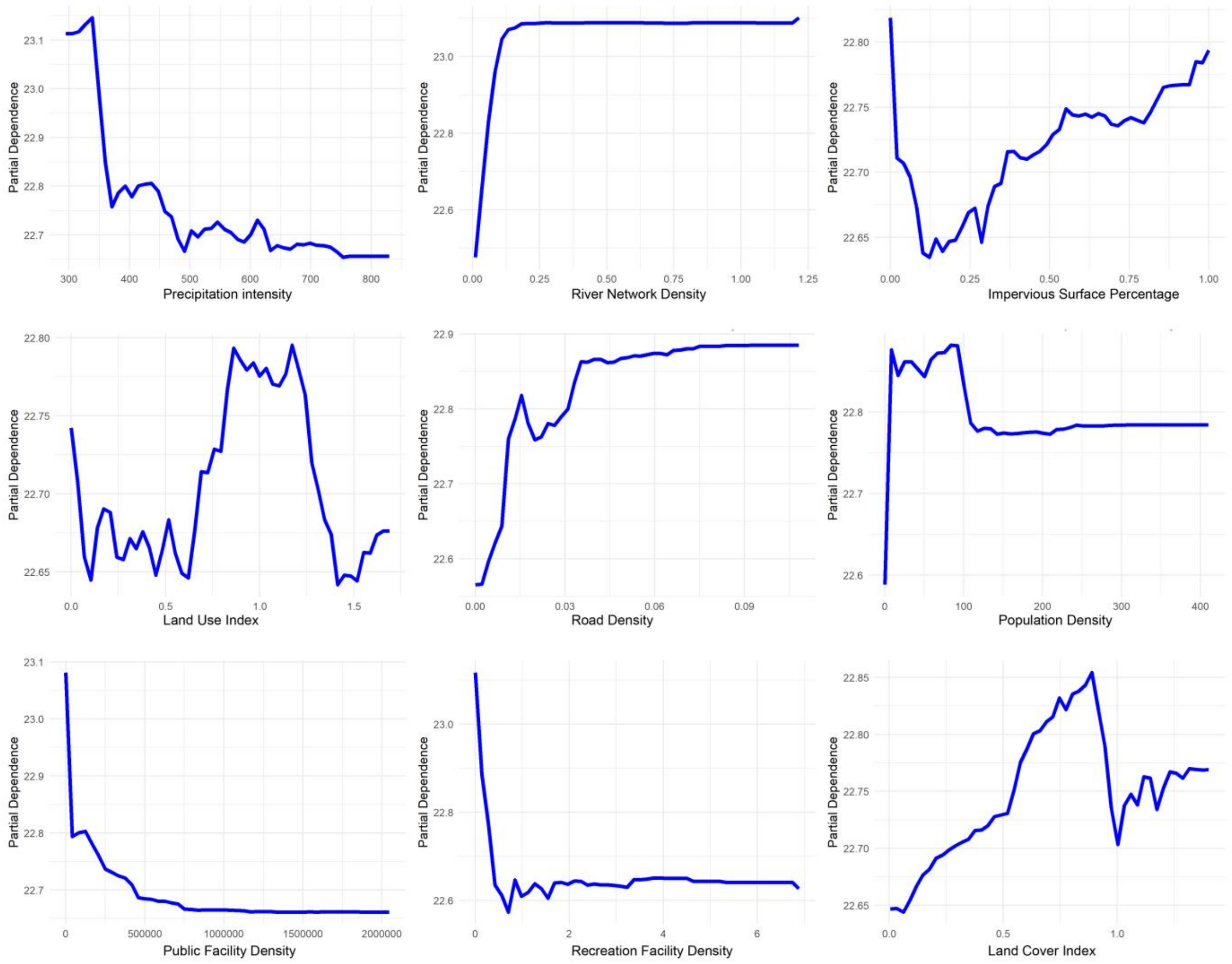

Figure S2. Partial dependence plots of the retained explanatory variables in the GWRF model.

To examine how local feature contributions vary across urbanization contexts, representative communities from high, medium, and low urbanization groups were further interpreted using local SHAP analyses (Figure S3). Force plots, interaction heatmaps, and SHAP dependence plots reveal distinct feature contributions, interactions, and nonlinear response patterns across the three urbanization levels.

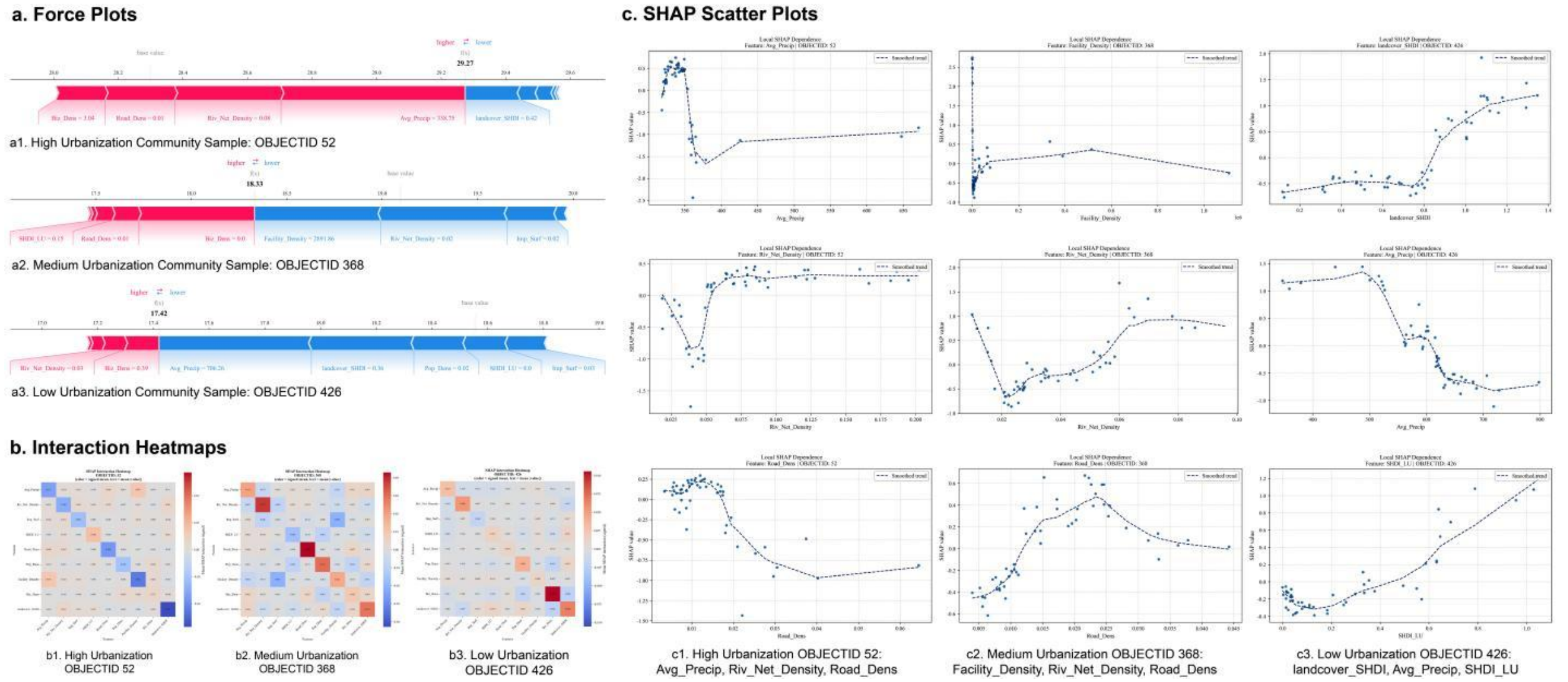

Figure S3. Local SHAP explanations across different urbanization levels. (a) Force plots for representative high-, medium-, and low-urbanization communities; (b) SHAP interaction heatmaps for the same communities; and (c) SHAP dependence plots showing nonlinear feature effects across urbanization levels.

**Urbanization stratification**

Communities were stratified according to the proportion of the urban population in the total population, a commonly used indicator of urbanization level. Following the Northam-based classification widely adopted in Chinese urbanization studies, communities with urban population ratios below 30% were classified as low urbanization, those between 30% and 70% as medium urbanization, and those above 70% as high urbanization. This stratification was used to organize and compare spatially heterogeneous Local Spatial TreeSHAP outputs across different urbanization contexts.

## 4 Field Survey

### Validation in Community Q

Qinglonghu Town is in a transitional zone between mountainous terrain and alluvial plains, representing a typical case of high urbanization with industrial agglomeration. The Dashi River and Ciwei River converge within the town, with Chongqing Reservoir situated between them. During the extreme rainfall event, Qinglonghu was among the 13 most severely affected towns in the district.

The town covers an area of 96 km², with a maximum accumulated rainfall of 612.8 mm. Runoff from shallow mountainous areas rapidly concentrated into large-scale flash floods, resulting in the disruption of 10 major roads, mobility difficulties in 26 villages, and severe damage in 6 villages. More than 20,000 residents were affected, with total economic losses approaching 1 billion RMB. Infrastructure damage included 41 roads, 23,000 meters of water supply pipelines, and 717,600 meters of drainage pipelines.

*"When rainfall reaches a certain level, we know we must prepare for evacuation; if it exceeds the threshold, evacuation must happen immediately. This is a scientific indicator."* (Interviewee 1)

*"Runoff from shallow mountainous areas concentrates into massive floods… water levels in the Dashi and Ciwei Rivers kept rising."* (Interviewee 2)

*"Village supermarkets opened their supplies for public use… we later required each village to maintain a 10-day emergency stock during flood seasons."* (Interviewees 1 & 3)

*"The old National Road 108 cannot be widened due to terrain constraints… temporary emergency roads were built using soil and gravel just to maintain basic access."* (Interviewees 1 & 4)

**Validation in Community F**

Township F is in the upstream region of the Dashi River basin and represents a typical low-density shallow mountainous area. The permanent population is approximately 4,000, and infrastructure provision is limited due to ecological conservation policies that restrict industrial development. During the 2023 extreme rainfall event, recorded precipitation exceeded 800 mm, and flood peaks surpassed design standards by more than twofold. All critical systems, including communication, electricity, roads, and water supply, were disrupted.

*"By around 11 p.m. that night, all stations had lost signal… once power was cut, they could only last a few hours."* (Interviewee 5)

*"We need distributed energy-supported communication stations in mountainous areas to avoid complete disconnection during disasters."* (Interviewee 6)

*"In emergencies, redundancy and standards are critical… but extreme events can still exceed design limits."* (Interviewees 7 & 8)

*"Much of the mountainous area is protected forest… development is restricted."* (Interviewee 7)

*"There are no regulating structures in the Dashi River… if we could reduce the flood peak, downstream impacts would be minimal."* (Interviewee 9)

**Validation in Community *S***

Town S is in the deep mountainous Juma River basin and is characterized by karst geomorphology, with widespread exposed rock and thin soil layers. This leads to poor water retention capacity and rapid runoff generation, making the area highly prone to flash floods. During the extreme rainfall event, total precipitation approached 700 mm over approximately 84 hours. Extensive damage occurred, including destruction of scenic areas, riverbanks, and transportation infrastructure.

*"Karst terrain has almost no inherent disaster resistance… it promotes rapid surface runoff and sediment transport."* (Interviewee 10)

*"The soil layer is extremely thin… landslides can occur, and recovery may take centuries."* (Interviewees 10 & 11)

*"Relocation can be at district level, township level, or nearby households… homestays can serve both tourism and emergency accommodation."* (Interviewee 11)

*"Relocating people to hotels creates significant financial burdens… we are still paying off those costs."* (Interviewee 12)